\documentclass{article} %
\usepackage{iclr2027_conference,times}

\usepackage{amsmath,amsfonts,bm}

\def\eqref#1{equation~\ref{#1}}

\def\1{\bm{1}}

\DeclareMathAlphabet{\mathsfit}{\encodingdefault}{\sfdefault}{m}{sl}
\SetMathAlphabet{\mathsfit}{bold}{\encodingdefault}{\sfdefault}{bx}{n}

\usepackage{xurl}      %
\usepackage{hyperref}
\usepackage{url}
\usepackage{xcolor}
\usepackage{graphicx}
\usepackage{booktabs}
\usepackage{multirow}
\usepackage{cleveref}   %
\providecommand{\dmodel}{d_{\mathrm{model}}}

\definecolor{cwrublue}{HTML}{003071}
\definecolor{myturquois}{HTML}{01AB8F}
\definecolor{mylightblue}{HTML}{1E90FF}
\hypersetup{colorlinks=true, linkcolor=mylightblue, citecolor=cwrublue, urlcolor=cwrublue}

\newcommand{\TMCR}{MMR}    %

\usepackage[most]{tcolorbox}
\definecolor{mypink}{HTML}{D6336C}
\definecolor{myorange}{HTML}{E97132}
\newtcolorbox{greybox}[1][]{float,title=#1,}
\newtcolorbox{bluebox}[1][]{float,title=#1,colback=myturquois!5,colframe=myturquois}
\newtcolorbox{pinkbox}[1][]{float,title=#1,colback=mypink!5,colframe=mypink}
\newtcolorbox{orangebox}[1][]{float,title=#1,colback=myorange!5,colframe=myorange}
\tcbset{aibox/.style={width=\linewidth,top=8pt,bottom=4pt,colback=blue!6!white,colframe=black,colbacktitle=black,enhanced,center,attach boxed title to top left={yshift=-0.1in,xshift=0.15in},boxed title style={boxrule=0pt,colframe=white,},}}
\newtcolorbox{AIbox}[2][]{aibox,title=#2,#1}

\usepackage{amsmath, amsthm}

\usepackage{enumitem}
\setlist[itemize]{
    leftmargin=*,       %
    labelsep=0.5em,     %
    nosep,              %
    itemindent=0pt,     %
    leftmargin=2.0em    %
}

\crefname{prop}{Proposition}{Propositions}
\Crefname{prop}{Proposition}{Propositions}

\title{How to Loop MoE: \\ Flatten the Experts, Untie the Attention}

\author{Shouren Wang$^{1, *}$,~Chuang Ma$^{2,3,*}$,~Mohsen~Hariri$^{1}$,~Debargha~Ganguly$^{1}$,~Wang~Yang$^{1}$ \\
\textbf{Xiaoqing~Tong$^{2}$,~Qianying~Liu$^{3}$,~Xiaotian~Han$^{1, \dagger}$,~Vipin~Chaudhary$^{1, \dagger}$} \\
$^{1}$Case Western Reserve University \\
\texttt{\{sxw992,dxg512,mxh1029,wxy320,vipin,xhan\}@case.edu} \\
$^{2}$Kyoto University \quad $^{3}$NII LLMC \\
\texttt{\{ma.chuang.52h,tong.xiaoqing.75d\}@st.kyoto-u.ac.jp} \quad \texttt{ying@nii.ac.jp} \\
$^{*}$Equal contribution \quad $^{\dagger}$Corresponding authors
}

\iclrfinalcopy %
\begin{document}

\maketitle
\lhead{Preprint}

\begin{abstract}
Looped Transformers reuse one block of layers several times: by spending extra computation they push a model of fixed size further, and so use its parameters more fully; while sparse mixture-of-experts (MoE) models activate only a few of many experts for each token. Looped MoE bridges these two design philosophies and gives MoE models new potential for better expert usage, but it raises a question: how to loop a MoE? We answer it with Foil. With the expert parameters and the expert compute per token held fixed, Foil (1) \emph{flattens} the experts, halving the expert layers, doubling the experts per layer and doubling the passes, so that every routing decision chooses from a larger pool, and (2) \emph{unties} the attention, giving each pass its own attention parameters while the experts and routers stay shared. Experiments show that Foil clearly outperforms the unflattened looped baseline: at 20B tokens every Foil model has lower pretraining loss than the baseline; at 100B tokens the loss improves monotonically with the degree of flattening, the most flattened Foil ending 0.012 nat below the baseline at equal parameters and compute, with downstream accuracy on par or better; untying the attention also yields more balanced and more confident routing at equal shape. Our ablations analyse why Foil works and turn the findings into design guidance for looped MoE: the returns of looping and of widening the expert layers amplify each other, routing confidence tracks healthy expert use better than load balance, and a sparse looped MoE should therefore use more experts per layer and more passes. Code and configurations are available at \url{https://github.com/SR-A-W/how-to-loop-moe}.
\end{abstract}

\begin{figure}[h]
\begin{center}
\includegraphics[width=0.8\linewidth]{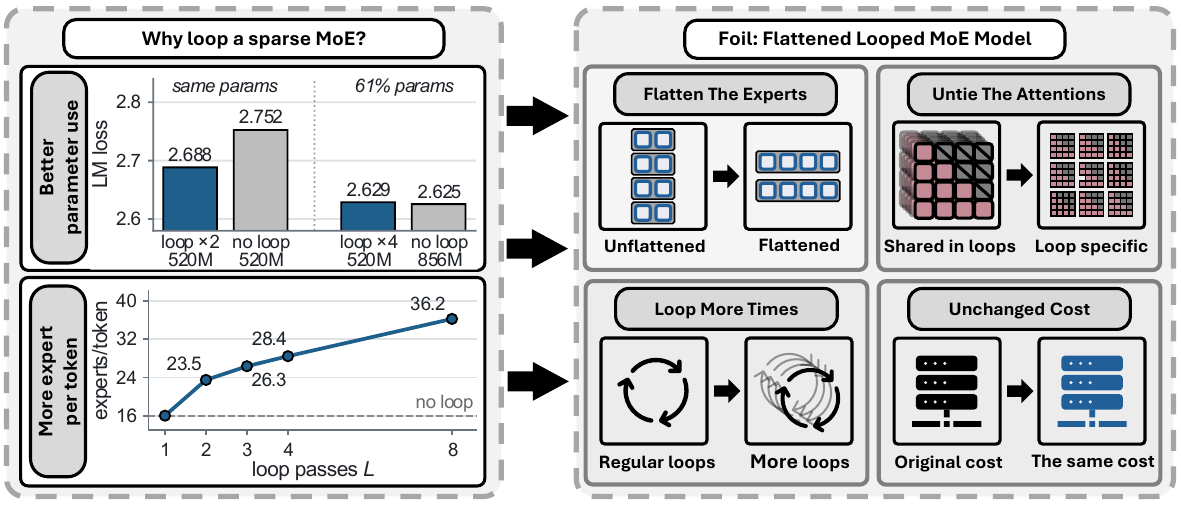}
\end{center}
\caption{Motivation of designing Foil. \textit{Left: why loop a sparse MoE}. (1) At equal parameters, looping lowers the loss, and matches non-looped models with much less parameters; (2) Tokens reaches more distinct experts the more passes it makes. \textit{Right: Foil's design}. (1) Flattens the experts, (2) Unties the attention. (3)Loops more times, (4)Keep parameters and compute unchanged.}
\label{fig:motivation}
\end{figure}

\providecommand{\Ereal}{E_{\mathrm{real}}}
\providecommand{\Ecomp}{E_{\mathrm{comp}}}
\providecommand{\Eeq}{E_{\mathrm{eq}}}

\section{Introduction}
\label{sec:intro}

Looped models apply one block of Transformer layers repeatedly to an evolving hidden state, so that the depth of computation is set by the number of passes rather than by the number of stored layers~\citep{dehghani2019universal}. A model of fixed size can thus be made stronger by computing more, which uses its parameters more fully: both theory and experiments show that looped Transformers suit computations that need many iterative steps~\citep{giannou2023looped, saunshi2025latent}, and looping has recently been scaled to large pretraining runs and to spending more computation at inference time~\citep{geiping2025scaling, zhu2025ouro}. As hardware compute grows far faster than memory capacity and bandwidth~\citep{gholami2024memorywall}, trading computation for stored parameters is increasingly attractive, and how best to loop a model has become an active question~\citep{prairie2026parcae, huang2026loopeddoneright}.

Sparse mixture-of-experts (MoE) models take a different route to efficiency: each layer holds many experts, but every token is routed to only a few of them, so that the parameter count far exceeds the computation per token; current large models hold hundreds of experts per layer and route each token to only a few~\citep{deepseekai2024v3, kimiteam2025k2}. 
The price is that a token sees only $k$ experts in each layer, and the more experts there are, the less often each of them is used. Whether the experts are actually used, and whether they specialise, has been a central question since the first sparse MoE models~\citep{fedus2022switch, zoph2022stmoe}, and a balanced load alone does not answer it~\citep{jin2026zloss}.

These two design philosophies are, however, naturally compatible. A routing decision exposes a token to only a few of the available experts; looping changes this: every pass gives the token a new routing decision in the same layer, so it can reach different experts, and different combinations of them, without storing any additional expert parameters. Our measurements make this potential concrete in three observations.
\emph{(1) Looping brings better performance or fewer parameters.} At equal parameters, looping twice lowers the loss by $0.064$~nat, and a looped model with far fewer parameters nearly matches a non-looped model with twice the layers by spending more computation per token (\Cref{fig:motivation}, top).
\emph{(2) Looping lets a sparse MoE use more of its experts.} The number of distinct experts a token reaches grows with the passes, from $16$ without looping to $36$ with eight passes (\Cref{fig:motivation}, bottom).
\emph{(3) Looping unlocks equivalent-parameter properties for MoE.} Because experts are called repeatedly, the equivalent number of experts and the number of possible routing combinations grow multiplicatively as the looped block is flattened, while the real experts and the compute stay fixed.

Earlier work has combined looping with experts~\citep{csordas2024moeut, chen2026loopmoe, jaggi2026tying, li2025megrez2}, but none has studied how the experts should be distributed over the looped block---how many experts a layer holds, how many layers the block has and how many times it is looped---when the expert parameters and the compute are fixed. Nor has it been asked which parameters a pass should own: with attention shared across passes, flattening discards the attention parameters of the layers it removes, whereas giving each pass its own attention keeps the parameter budget and lets successive passes process the shared experts' inputs differently, at no extra compute.
This raises our question: \textbf{how to loop a MoE when its parameters and compute are fixed?} We answer it with Foil. Foil (1)~\emph{flattens} the experts, scaling down the number of layers in the looped block while scaling up the experts per layer and the number of passes in proportion, so that every routing decision chooses from a larger pool, and (2)~\emph{unties} the attention, giving each pass its own attention parameters while the experts and routers stay shared across passes. Our contributions are:

\begin{itemize}
\item We propose \textbf{Foil}, which flattens the experts and unties the attention of a looped MoE while holding the parameters and the compute per token fixed (\Cref{sec:method}).

\item We validate Foil in 20B-token pretraining and 100B-token continued training: it clearly lowers the pretraining loss of the unflattened baseline, matches or exceeds it downstream, and untying the attention yields healthier routing than tying it at the same shape (\Cref{sec:experiments}).

\item Through systematic ablations we characterise several phenomena of looped MoE and distil a design suggestion: a sparse looped MoE should use appropriately more experts per layer and more passes (\Cref{sec:ablation}).
\end{itemize}

\section{Methodology}
\label{sec:method}

\begin{figure}[t]
\begin{center}
\includegraphics[width=\linewidth]{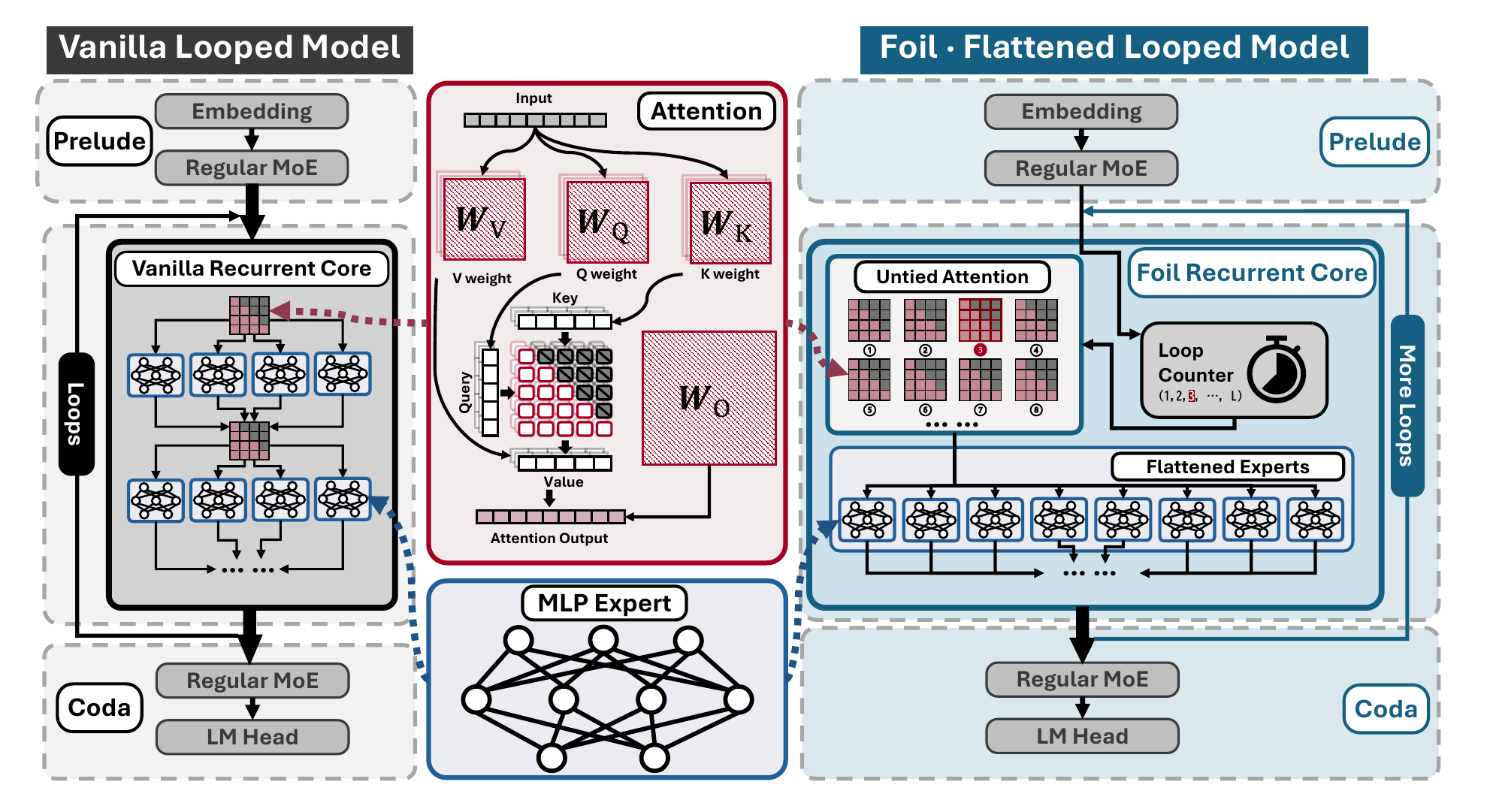}
\end{center}
\caption{Vanilla looped MoE versus \emph{Foil}. Both models share the same skeleton: a prelude (embedding and one regular MoE layer), a recurrent core applied for several passes, and a coda (one regular MoE layer and the LM head). The vanilla core ties both attention and experts across passes. Foil flattens the core---it keeps the total number of experts fixed while using fewer layers, more experts per layer and more passes---and unties the attention: the experts are shared across passes, but every pass has its own attention set, selected by the pass index. Both models call the same number of experts per token. When the core has several layers, every layer carries one attention set per pass; the ellipsis marks the remaining sets. Red: attention; blue: experts.}
\label{fig:architecture}
\end{figure}

\subsection{Notation and accounting}
\label{sec:method-notation}

We first fix the notation for the looped block and count what it stores and computes; \Cref{fig:architecture} shows the vanilla looped MoE and Foil side by side, and \Cref{sec:method-foil} defines Foil.

\paragraph{Resource and routing accounting.}
Consider a block with $D$ separate banks of $E$ experts, each containing $p_{\mathrm{exp}}$ parameters, traversed $L$ times. Assume unrestricted top-$k$ routing with $1\leq k\leq E$, exactly $k$ distinct experts executed per visit, and no dropped assignments. Then $\Ereal=E\times D$, $D_{\mathrm{eff}}=D\times L$, $\Eeq=E\times D\times L$, $\Ecomp=k\times D\times L$ and $P_{\mathrm{experts}}=E\times D\times p_{\mathrm{exp}}$ (router scoring, below $1\%$ of the expert compute for the most flattened shape, is not counted; Appendix~\ref{sec:app-theory}).

Appendix~\ref{sec:app-theory} gives a rigorous formulation of the looped block, of these identities and of the bounds on expert coverage.

\subsection{Foil: flatten the experts, untie the attention}
\label{sec:method-foil}

We build on the looped skeleton of Huginn~\citep{geiping2025scaling}: token embedding, a \emph{prelude} of one ordinary MoE layer, a looped block of $D$ layers applied $L$ times, a \emph{coda} of one ordinary MoE layer, and the output layer. The prelude and coda have $8$ experts each, run once and are never flattened, so all configurations differ only in the looped block, whose shape we write as $(E,D,L)$ (\Cref{fig:architecture}).

\paragraph{Flattening enlarges the routing pool at fixed expert budget and compute.}
One flattening step maps $(E,D,L)$ to $(2E,D/2,2L)$: half the layers, twice the experts per layer, twice the passes. It keeps the expert parameters ($\Ereal=E\times D$), the expert calls per token ($\Ecomp=k\times D\times L$) and the effective depth ($D_{\mathrm{eff}}=D\times L$) fixed, and enlarges the pool $E$ of every routing decision and the equivalent expert count $\Eeq=E\times D\times L$, the number of experts in the non-looped model obtained by unrolling the passes. Three steps from $(8,8,2)$ give $(16,4,4)$, $(32,2,8)$ and $(64,1,16)$, raising $\Eeq$ from $128$ to $1024$.

\paragraph{Untying the attention restores discarded parameters at no extra compute.}
A conventional looped model shares the whole block across passes, attention included, so flattening also removes attention sets: the looped block of $(8,8,2)$ holds eight, that of $(64,1,16)$ a single one reused on every pass, while the experts are untouched. We therefore share the experts and routers across passes but give every pass its own attention, $DL=16$ sets along the whole sequence, with no pass embedding (\Cref{fig:architecture}, right). Untying adds no computation; it only restores the attention parameters that sharing discards. We call the flattened models with shared attention \emph{proto-Foil} and those with untied attention \emph{Foil}.

We propose three Foils: Foil-1 $(64,1,16)$, Foil-2 $(32,2,8)$ and Foil-3 $(16,4,4)$. The baseline, Base $(8,8,2)$, is the looped model with untied attention and differs from the Foils only in shape; the controls, Base-tied and proto-Foil-3, -2, -1, share the attention. Base and the Foils have $553.7$M parameters each, the shared-attention models $490.8$--$520.2$M. The further models of the ablations (\Cref{sec:ablation}) are named by their shape. All models route each token to the $k=2$ most probable experts of a layer under a linear router with softmax and renormalised weights, with SwiGLU experts (all settings and parameter counts in \Cref{tab:app-hparams}).
\subsection{Three metrics describe how the experts are used}
\label{sec:method-metrics}

We measure expert use on the recurrent core over $255{,}500$ probe tokens; $S_{t,d}^{(\ell)}$ are the $k$ experts token $t$ selects at layer $d$ and pass $\ell$, and $p_{t,(1)}\geq\cdots\geq p_{t,(E)}$ its sorted router probabilities.

\paragraph{Distinct experts reached per token ($U_t$).}
\begin{equation*}
U_t=\sum_{d=1}^{D}\Bigl|\,\bigcup_{\ell=1}^{L}S_{t,d}^{(\ell)}\Bigr|\;\leq\;D\min(E,kL)
\end{equation*}
counts the distinct experts token $t$ reaches over its passes; its mean, as a fraction of the bound, shows how many experts looping lets a token use.

\paragraph{Load balance ($B_2$).}
\begin{equation*}
B_2=\frac{1}{E\sum_{i=1}^{E}q_i^{2}}=\frac{N_2}{E},
\end{equation*}
where $q_i$ is the share of routing requests expert $i$ of a layer receives, passes pooled, and $N_2$ the effective number of experts~\citep[Eq.~S10]{wu2026reba}. This is Jain's fairness index~\citep{jain1984fairness}: it lies in $[k/E,1]$, equals $1$ for even load and shows whether the load concentrates on a few experts; model values pool $N_2$ over layers.

\paragraph{Routing confidence: the median margin ratio (\TMCR{}).}
\begin{equation*}
\mathrm{\TMCR{}}=\frac{p_{t,(1)}}{p_{t,(E/2)}}=\exp\bigl(z_{t,(1)}-z_{t,(E/2)}\bigr)
\end{equation*}
compares the router's first choice with the median-ranked expert of the whole pool ($z$: router logits), a margin taken against the median expert rather than within or at the edge of the selected set. It equals $1$ for an indifferent router at any $E$, is averaged geometrically over decisions, and targets balanced load with indifferent routing~\citep{jin2026zloss}.

\emph{The metrics are read together, relative to themselves, or at equal shape.} Router scores are not rescaled, and both $B_2$ and \TMCR{} depend on $E$; only same-direction changes of both are read as healthier routing. A traffic-matched masking test checks whether rarely used experts are dispensable (Appendix~\ref{sec:app-masking}); related metrics are in Appendix~\ref{sec:app-metrics}.

\section{Experiments}
\label{sec:experiments}

Every model is trained on the same data in the same order, from the same initialisation seed and under the same learning-rate schedule, for 20B tokens (training details in Appendix~\ref{sec:app-training}); this is 35--40 tokens per parameter for the eight compared models, enough by the Chinchilla ratio~\citep{hoffmann2022chinchilla}, and we scale the main models up to 100B tokens. The router applies a softmax at temperature $1$ over the experts of a layer and selects the top two, without noise or bias terms.

Downstream, the main text reports three zero-shot tasks, one representative for each of word prediction (LAMBADA, standard split), sentence continuation (HellaSwag) and coreference resolution (XWinograd, English). We report (1) language-modelling loss at the end of training, compared between models as paired differences with standard errors, (2) downstream task results, and (3) routing metrics ($B_2$ and \TMCR{}). Full results are provided in Appendix~\ref{sec:app-full}, including a seed-change experiment that supports the robustness of the results.

\emph{The runs of this study consumed approximately \textbf{12.6k NVIDIA H200 GPU-hours} (about 18.2k including earlier control and failed runs); each run used at most one node with eight H200 GPUs.}

\subsection{Flattening with more passes raises the ceiling of model ability}
\label{sec:exp-flatten}

Holding the real expert count $\Ereal=64$ and the expert calls per token $\Ecomp=32$ fixed, we flatten Base step by step, halving the layers of the recurrent core, doubling the experts per layer and doubling the passes, which gives Foil-3, Foil-2 and Foil-1; every pass keeps its own attention.

\begin{figure}[t]
\begin{center}
\includegraphics[width=\linewidth]{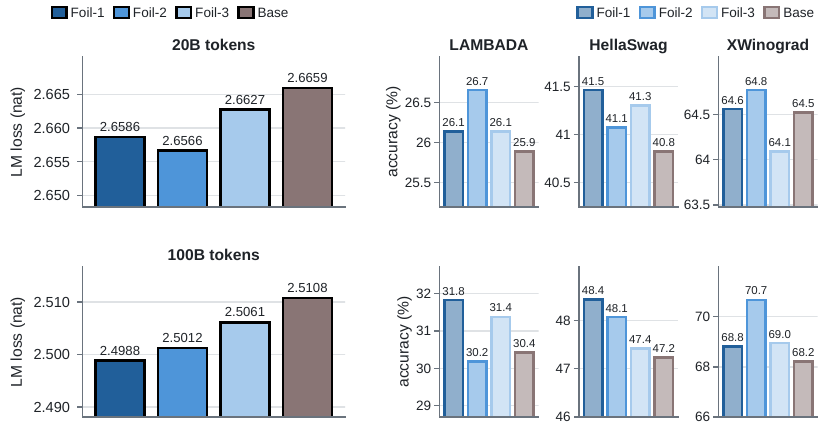}
\end{center}
\caption{Flattening with untied attention: loss and downstream accuracy of the Foils and Base. Top row: 20B tokens; bottom row: after continued training to 100B tokens. In each row, solid bars on the left show the final loss (lower is better) and tinted bars on the right the zero-shot accuracy on the three representative tasks (higher is better), each panel with its own vertical axis; bars from left to right: Foil-1, Foil-2, Foil-3, Base. At 20B \textbf{all three Foils are below Base}; at 100B \textbf{the loss falls monotonically with flattening} and Foil-1 ends \textbf{$0.012$~nat below Base}. Downstream accuracy is on par or slightly better; further tasks are in Appendix~\ref{sec:app-eval}.}
\label{fig:main}
\end{figure}

Flattening improves the model at both training lengths (\Cref{fig:main}):
\begin{itemize}
\item \textbf{At 20B tokens}, all three Foils reach a lower loss than Base. The loss falls through the second flattening step and then levels off: Foil-1 ends $0.007$~nat below Base, slightly above Foil-2. On the three representative tasks, the Foils are at or above Base, within their standard errors.
\item \textbf{At 100B tokens}, every flattening step lowers the loss and Foil-1 is the strongest, $0.012$~nat below Base. On the three representative tasks, all three Foils score slightly above Base, by one to two standard errors.
\end{itemize}
The 100B results confirm the 20B findings and enlarge them: the gain of the flattest shape grows with training, while Foil-2's stays put. Further downstream results are in \Cref{tab:app-eval-0shot}. The remaining experiments use 20B tokens.

\subsection{Untied attention unlocks the flattened model's potential}
\label{sec:exp-untied}

At each of the four shapes we compare a pair of models that differ only in whether the attention is tied across passes: Base-tied and proto-Foil-3/2/1 share one attention set over all passes, whereas Base and Foil-3/2/1 keep one per pass (\Cref{tab:app-hparams}). The tied models follow the classic looped design and serve as controls, but flattening discards their attention parameters ($520.2$M to $490.8$M), whereas Base and the Foils all have $553.7$M at equal compute.

\begin{figure}[t]
\begin{center}
\includegraphics[width=\linewidth]{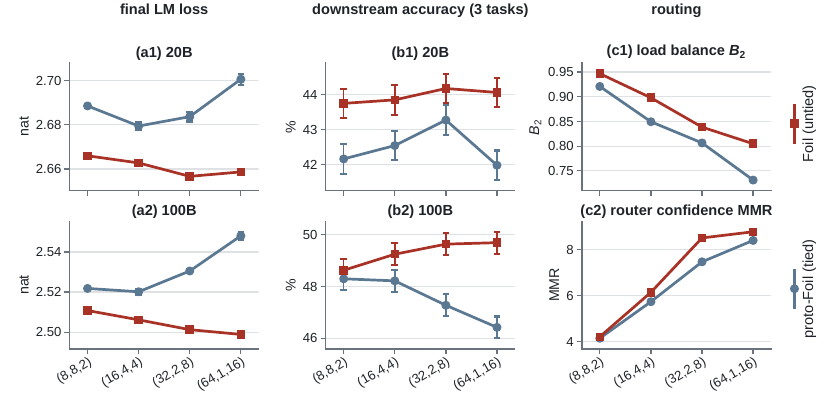}
\end{center}
\caption{Tied versus untied attention at equal shape: loss (a), downstream accuracy (b) and routing (c). Horizontal axis: flattening steps from $(8,8,2)$ to $(64,1,16)$; each step to the right doubles $E$ and $L$ and halves $D$. Red squares: untied attention (Base and Foil-3, -2, -1); grey-blue circles: tied attention (Base-tied and proto-Foil-3, -2, -1). (a1,~a2)~Final loss at 20B and 100B tokens; error bars are twice the standard error of the paired difference from the $(8,8,2)$ model of the same line. (b1,~b2)~Mean zero-shot accuracy over the three representative tasks, $\pm1$ standard error. (c1,~c2)~Load balance $B_2$ and routing confidence \TMCR{} of the recurrent core at 20B tokens, from a single-seed probe without error bars; since both depend on $E$, only the two points at the same shape are compared. With tied attention the loss rises again after the first flattening step, while with untied attention it levels off at 20B and keeps falling at 100B; the gap is largest for the most flattened pair ($0.049$~nat at 100B). At every shape the untied model is both more balanced and more confident.}
\label{fig:untied}
\end{figure}

\paragraph{Untying lowers the loss at every shape, more so when flatter.}
At 20B tokens, the untied model is better in all four pairs, most of all when fully flattened: Foil-1 is $0.042$~nat below proto-Foil-1 (\Cref{fig:untied}, a1--a2). After continued training to 100B tokens, the gap grows steadily with flattening, to $0.049$~nat for the Foil-1 pair.
\paragraph{Downstream, the untied model scores higher in every pair.}
On the mean of the three representative tasks, the untied model is ahead in all four pairs at both token budgets; at 20B the difference exceeds two standard errors in every pair except the Foil-2 pair, and at 100B it widens with flattening and exceeds two standard errors in the two most flattened pairs, reaching $3.3$ points for Foil-1 over proto-Foil-1 (\Cref{fig:untied}, b1--b2). Further downstream results are in \Cref{tab:app-eval-0shot}.

\paragraph{At equal shape, untied attention routes more evenly and more confidently.}
Reading the two routing metrics together and only at equal shape (\Cref{sec:method-metrics}), the untied model has both a higher load balance $B_2$ and a higher routing confidence \TMCR{} than its tied counterpart in all four pairs, with the smallest gap at the Base pair (\Cref{fig:untied}, c1--c2). The two metrics move in the same direction, which we read as healthier routing.

\begin{table}[t]
\caption{Final loss (nat) of the tied-attention models at 20B tokens and its decrease from the model with half the passes or half the experts per layer (standard errors $0.0006$--$0.0012$); --: not applicable.}
\label{tab:abl-gains}
\centering
\scriptsize
\setlength{\tabcolsep}{3pt}
\setlength{\aboverulesep}{0.2ex}\setlength{\belowrulesep}{0.3ex}
\begin{tabular*}{\linewidth}{@{\hspace{10pt}\extracolsep{\fill}}cccccccc@{\hspace{4pt}}}
\toprule
& & & & & & \multicolumn{2}{c}{\footnotesize $\Delta$ loss vs.\ model with} \\
\cmidrule(lr){7-8}
\footnotesize $D$ & \footnotesize $E$ & \footnotesize $L$ & \footnotesize $\Ereal$ & \footnotesize $k/E$ & \footnotesize Final loss (nat) & \footnotesize half the passes & \footnotesize half the experts per layer \\
\midrule
\multirow{6}{*}{$4$} & \multirow{3}{*}{$8$} & $2$ & 32 & $1/4$ & $2.7874$ & -- & -- \\
 &  & $4$ & 32 & $1/4$ & $2.7196$ & $0.068$ & -- \\
 &  & $8$ & 32 & $1/4$ & $2.6875$ & $0.032$ & -- \\
\cmidrule[0.3pt](lr){2-8}
 & \multirow{3}{*}{$16$} & $2$ & 64 & $1/8$ & $2.7490$ & -- & $0.038$ \\
 &  & $4$ & 64 & $1/8$ & $2.6793$ & $0.070$ & $0.040$ \\
 &  & $8$ & 64 & $1/8$ & $2.6384$ & $0.041$ & $0.049$ \\
\midrule
\multirow{6}{*}{$8$} & \multirow{2}{*}{$4$} & $2$ & 32 & $1/2$ & $2.7277$ & -- & -- \\
 &  & $4$ & 32 & $1/2$ & $2.6747$ & $0.053$ & -- \\
\cmidrule[0.3pt](lr){2-8}
 & \multirow{3}{*}{$8$} & $2$ & 64 & $1/4$ & $2.6884$ & -- & $0.039$ \\
 &  & $4$ & 64 & $1/4$ & $2.6287$ & $0.060$ & $0.046$ \\
 &  & $8$ & 64 & $1/4$ & $2.5961$ & $0.033$ & -- \\
\cmidrule[0.3pt](lr){2-8}
 & $16$ & $2$ & 128 & $1/8$ & $2.6414$ & -- & $0.047$ \\
\bottomrule
\end{tabular*}
\end{table}

\section{Ablation Studies}
\label{sec:ablation}

We run the ablations at 20B tokens, where \Cref{sec:experiments} showed that the trends agree with those at 100B tokens and that changing the initialisation seed alone barely moves the loss. Unless stated otherwise, the models in this section tie the attention across loop passes, so that adding passes adds no parameters (untying it keeps the gains of flattening, \Cref{sec:exp-untied}), and are named by their shape $(E,D,L)$. A \emph{gain} is a decrease of the final loss (nat), paired as in \Cref{sec:experiments}.

\subsection{Wider, sparser layers slow the diminishing returns of looping}
\label{sec:abl-loops}

We add passes, from two to four and from four to eight, to four layer configurations $(E,D)$ and measure the gain of each step at fixed $(E,D)$: $(8,4)$ and $(4,8)$ with $\Ereal=32$, and $(16,4)$ and $(8,8)$ with $\Ereal=64$.
\paragraph{Wider, sparser layers gain more from additional passes.}
From four to eight passes (\Cref{tab:abl-gains}, passes column), $(16,4)$ gains $0.041$~nat, against at most $0.033$ for $(8,4)$, which has half its real experts, and $(8,8)$, which has twice its layers. From two to four passes (\Cref{tab:abl-gains}), $(16,4)$ gains more than $(8,8)$ and is on par with $(8,4)$. At $\Ereal=32$, the sparse $(8,4)$ gains clearly more from two to four passes than the half-active $(4,8)$, although $(4,8)$ spends twice the expert compute per pass. Since each comparison changes more than one quantity, we conclude only that wider, sparser layers make the returns of looping decline more slowly, without attributing this to a single cause.

\subsection{More passes enlarge the gain from widening}
\label{sec:abl-widen}

Here \emph{widening} doubles the experts per layer $E$ at fixed $D$ and $L$, so the expert calls per token $\Ecomp$ stay the same while the real experts $\Ereal$ double.

\paragraph{Widening and looping amplify each other.}
Widening $(8,4,L)$ to $(16,4,L)$ gains more the more passes the model makes, from $0.038$~nat at $L=2$ to $0.049$ at $L=8$, and widening $(4,8,L)$ to $(8,8,L)$ likewise gains more at $L=4$ than at $L=2$: the more passes, the more widening pays. For each $2\times2$ block we take the gain of doing both minus the gains of widening alone and of looping alone; of the three such interactions, two are clearly positive and one is on par with zero, and none is negative (\Cref{tab:abl-gains}, experts column, as its increase with $L$).

\paragraph{Widening shows diminishing returns in the number of experts added.}
Widening once more, from $(8,8,2)$ to $(16,8,2)$, gains only $1.2$ times as much as widening from $(4,8,2)$ to $(8,8,2)$, although it adds eight experts per layer instead of four (\Cref{tab:abl-gains}, experts column); at two passes the added width is used less, in line with the finding above that widening pays more with more passes. Since our design cannot widen a layer at fixed $\Ereal$, part of this decline may come from the change in $\Ereal$.

\begin{AIbox}{Takeaway 1}
Widening the expert layers and looping more are complementary: each enlarges the other's gain. Under a fixed expert-parameter and per-token compute budget, this is exactly what flattening does: wider, sparser layers looped more often.
\end{AIbox}

\begin{figure}[h]
\begin{center}
\includegraphics[width=\linewidth]{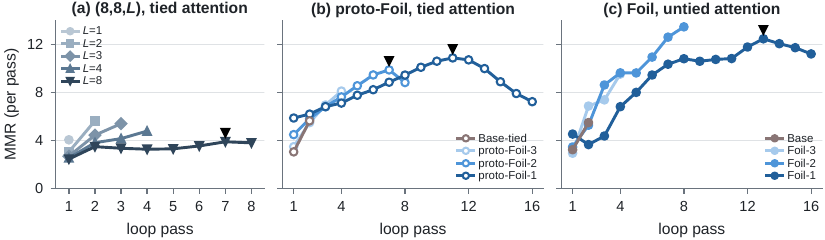}
\end{center}
\caption{Routing confidence (\TMCR{}) per loop pass at 20B tokens (geometric mean over the core layers). (a)~At fixed $(E,D)=(8,8)$, more passes lower the confidence of every pass, and beyond $L=2$ the model-level \TMCR{} falls from $4.15$ to $3.36$; $(8,8,8)$ peaks at pass~7. (b,~c)~Along the flattening sequence, confidence generally rises over the passes; proto-Foil-2, proto-Foil-1 and Foil-1 reach a peak (black triangles) and then fall, whereas Foil-2, the Foil with the lowest loss at 20B, has no peak. After a peak, further flattening or more passes very likely gain least (\Cref{sec:abl-peak}).}
\label{fig:abl-tmcr}
\end{figure}

\subsection{Load balance alone does not indicate healthy specialisation}
\label{sec:abl-balance}

Load balance is the classic measure of expert utilisation, but it has been questioned: a balanced load can hide a router that has no preference among the experts, which motivated measuring routing confidence as well~\citep{jin2026zloss}.

\paragraph{With untied attention, flattening routes more confidently, less evenly, and reaches a lower loss.}
Along the flattening sequence with untied attention, $B_2$ falls steadily from the unflattened model (Base) to the most flattened one (Foil-1), while \TMCR{} rises (\Cref{fig:untied}, c1--c2). Following the readings of \Cref{sec:method-metrics}, we read these only as trends of each metric, not as absolute comparisons across different $E$. Yet the most flattened model reaches a lower loss than the unflattened one (\Cref{fig:main}): the lower balance does not come with a worse model.

\paragraph{The least-used experts are not the least useful.}
Masking the experts that carry the least $10\%$ of the routing traffic and comparing with traffic-matched random groups (Appendix~\ref{sec:app-masking}), the least-used group raises the loss more than the random mean in seven of the eight models compared in \Cref{sec:experiments} (Base-tied, proto-Foil-3/2/1, Base, Foil-3/2/1), and in none does it fall below the range of the random groups; the exception, proto-Foil-2, is on par with the random median (\Cref{tab:app-masking-ratio} and \Cref{fig:app-masking-10}). For sparse MoE, load balance alone is therefore not a suitable indicator of healthy expert specialisation.

\subsection{Routing confidence and its per-pass peak index the looping gain}
\label{sec:abl-peak}

In \Cref{sec:exp-untied}, the \TMCR{} difference between Foil and proto-Foil is largest for the Foil-2 pair ($14\%$; \Cref{fig:untied}, c2), and Foil-2 has the lowest 20B loss of the Foil family including Base (\Cref{fig:main}). This led us to examine \TMCR{} pass by pass.

\paragraph{Routing confidence and the gain per pass decline together.}
In the $(8,8,L)$ series, where every model has eight experts per layer and absolute values are therefore comparable, the model-level \TMCR{} falls from $4.15$ to $3.36$ and the average gain per pass over the non-looped $(8,8,1)$ from $0.064$ to $0.022$~nat as $L$ goes from $2$ to $8$; the two decline together at every step, and $(8,8,8)$ is lowest in both (\Cref{tab:app-per-pass}). We state only that the two move in the same direction, not that one causes the other.

\paragraph{After a peak, further flattening or more passes very likely bring the smallest or no gain.}
In many models \TMCR{} rises with the pass index up to some pass and falls afterwards (\Cref{fig:abl-tmcr}): $(8,8,8)$, proto-Foil-2, proto-Foil-1 and Foil-1 peak, whereas the Base and Base-tied pair, the Foil-3 and proto-Foil-3 pair, Foil-2 and the $(8,8,L)$ models with $L\le4$ are most confident on their last pass. After its peak, flattening proto-Foil-2 further to proto-Foil-1 raises the loss by $0.017$~nat (\Cref{fig:untied}, a1); Foil-1, which peaks, is $0.0020$~nat above Foil-2 at 20B, about three standard errors (\Cref{fig:main}); and $(8,8,8)$ has the smallest average gain per pass of its series. Conversely, Foil-2, which has no peak, is the Foil with the lowest loss. These are few cases, so we state the relation as very likely rather than as a rule.

\begin{AIbox}{Takeaway 2}
Within a range, routing confidence (\TMCR{}) is a diagnostic of router differentiation beyond load balance; its per-pass peak signals that the gains from looping are close to exhausted, a practical guide for designing looped models.
\end{AIbox}

\section{Related Work}
\label{sec:related}

\paragraph{Looped Transformers.}
Looped models apply the same layers repeatedly to an evolving hidden state, trading computation for depth and reuse for parameters, from the Universal Transformer to latent reasoning and scaling laws for recurrent depth~\citep{dehghani2019universal, giannou2023looped, saunshi2025latent, geiping2025scaling, zhu2025ouro, bae2025relaxed, mcleish2026retrofitted, schwethelm2026isodepth, prairie2026parcae}. Their looped blocks are dense; ours is a sparse MoE, whose experts and attention must be arranged over layers and passes.

\paragraph{Sparse mixture-of-experts.}
Sparse MoE models activate only a few experts per token and thus grow their parameters far beyond their computation, and they now underlie most large language models~\citep{fedus2022switch, lepikhin2021gshard, zoph2022stmoe, jiang2024mixtral, deepseekai2024v3, qwenteam2025qwen3, muennighoff2025olmoe, kimiteam2025k2, kimiteam2026k3, glm5team2026, minimax2026m2}; the granularity of the experts is a further design dimension, splitting experts into smaller ones and activating more of them~\citep{dai2024deepseekmoe, he2024peer, ludziejewski2024scaling}. Earlier work combines looping with experts by turning the layers of a looped Transformer into shared mixtures of experts or by looping a whole MoE block~\citep{csordas2024moeut, chen2026loopmoe}; MoEUT ablates how many distinct consecutive layers form its repeated group and uses two for its smaller models. Other work ties experts across neighbouring layers~\citep{jaggi2026tying, tan2025rexmoe, chen2026moue}: MoRE lets adjacent layers share one larger expert pool, each layer keeping its own router~\citep{qiu2026more}, and Megrez2 is similar~\citep{li2025megrez2}. In dense models, One Wide FFN shares one widened feed-forward layer across the encoder layers while keeping per-layer attention~\citep{pires2023onewide}, a dense counterpart of flattening with untied attention. Each proposes one form of sharing, but none compares, at fixed expert parameters and compute, the layout $(E,D,L)$ of the looped block and whether its attention is shared across passes; \citet{jaggi2026tying} and Megrez2 share experts with per-layer attention, as we do, but start from compressing deep models rather than varying the layout of a looped block.

\paragraph{Expert utilisation.}
Load balance is usually maintained by auxiliary losses, capacity limits or bias-based balancing without an auxiliary loss~\citep{fedus2022switch, zoph2022stmoe, wang2024auxfree}; beyond imbalance, experts can degrade through representation collapse or a router without preference~\citep{chi2022representation, jin2026zloss}, and recent routing and balancing methods build on the same score-distribution and effective-count views~\citep{shahout2025laser, nguyen2026libmoe, wu2026reba}. We keep a standard balance metric, add \TMCR{}, which measures confidence against the median of all experts, and a traffic-matched masking test, and interpret them only through the three readings of \Cref{sec:method-metrics}.
\section{Conclusion}
\label{sec:conclusion}

How should a MoE be looped? We answer with Foil, which flattens the experts into fewer, wider layers looped more often and unties the attention across passes. At equal parameters and compute, Foil outperforms the unflattened looped baseline, its loss improves monotonically with flattening at 100B tokens, and at every shape it beats the attention-sharing proto-Foil in loss, downstream accuracy and, at 20B, routing balance and confidence. Our ablations show that widening and looping amplify each other, that load balance alone does not indicate healthy specialisation while \TMCR{} moves with the looping gain, and that its per-pass peak marks where flattening gains least. Limitations and future work are discussed in Appendix~\ref{sec:app-limitations}.

\section*{Acknowledgments}
This work was supported in part by NSF awards 2117439 and 2112606.

We thank Hongye Jin, whose early discussions motivated and inspired this project, for generously sharing insights and expertise throughout.

\bibliography{reference}
\bibliographystyle{iclr2027_conference}
\appendix

\section{Formal setup and resource accounting}
\label{sec:app-theory}
This section states the looped block of \Cref{sec:method-notation} formally and derives the identities and bounds used there.

\paragraph{Problem formulation.}
We seek to improve sparse mixture-of-experts language models by reorganising how their parameters are stored and reused. As in \Cref{sec:method-notation}, $E$ is the number of experts in each physical layer of the looped block, $D$ the number of layers in the block, $L$ the number of passes through it, and $k$ the number of experts selected per token at each layer visit. Given a token sequence $x_{1:T}$, the prediction objective is the next-token negative log-likelihood
\begin{equation}
\mathcal{L}_{\mathrm{LM}}=-\mathbb{E}_{x_{1:T}}\Bigl[\frac{1}{T}\sum_{t=1}^{T}\log p_\theta(x_t\mid x_{<t})\Bigr].
\end{equation}
The goal is to lower this loss while controlling the stored expert parameters and the selected-expert computation per token; the quantities below are those that Foil (\Cref{sec:method-foil}) holds fixed or enlarges.

\paragraph{The looped block and attention sharing.}
Let $H_d^{(\ell)}$ be the hidden states of the sequence after physical layer $d$ on pass $\ell$, with $H_0^{(1)}$ the output of the prelude. Write $\mathcal{A}_d^{(\ell)}$ for the attention sublayer of layer $d$ on pass $\ell$ and $\mathcal{M}_d$ for its MoE sublayer, each including normalisation and the residual connection. The looped block computes
\begin{equation}
H_d^{(\ell)}=\mathcal{M}_d\bigl(\mathcal{A}_d^{(\ell)}(H_{d-1}^{(\ell)})\bigr),
\qquad H_0^{(\ell)}=H_D^{(\ell-1)}\quad(\ell>1),
\label{eq:loop-composition}
\end{equation}
and passes $H_D^{(L)}$ to the coda; attention, routing and expert outputs are recomputed on every visit. The experts and router of $\mathcal{M}_d$ are shared across passes, but their inputs change with $\ell$, so sharing does not force the same expert selections on different passes. Tied attention imposes $\mathcal{A}_d^{(\ell)}=\mathcal{A}_d$ for all $\ell$; Foil gives every pass its own attention map. At fixed $(E,D,L)$, every tied model is recovered from an untied one by setting its per-pass attention weights equal, so the tied function class is contained in the untied one; this guarantees neither strict inclusion nor better optimisation. Different flattened shapes impose different sharing constraints, so the containment does not extend across shapes.

\paragraph{Routing.}
Let $u_{t,d}^{(\ell)}$ be the normalised input of token $t$ to the MoE sublayer of layer $d$ on pass $\ell$, $f_{d,e}$ the $e$-th expert of that layer, and $g_d$ and $w_{d,e}$ its router scores and expert combination weights. The selected experts and their combined output are
\begin{align}
S_{t,d}^{(\ell)}&=\operatorname{TopK}\bigl(g_d(u_{t,d}^{(\ell)}),k\bigr),\\
y_{t,d}^{(\ell)}&=\sum_{e\in S_{t,d}^{(\ell)}}w_{d,e}(u_{t,d}^{(\ell)})\,f_{d,e}(u_{t,d}^{(\ell)}),
\end{align}
and $\mathcal{M}_d$ adds $y_{t,d}^{(\ell)}$ to the residual stream.

\paragraph{Resource and routing identities.}
Let each expert contain $p_{\mathrm{exp}}$ parameters, and assume unrestricted top-$k$ routing with $1\leq k\leq E$, exactly $k$ distinct experts executed per visit, and no dropped assignments. Counting each physical expert once for storage, each layer visit once for effective depth, and each selected expert once per call gives the identities of \Cref{sec:method-notation},
\begin{align*}
\Ereal&=E\times D, & P_{\mathrm{experts}}&=E\times D\times p_{\mathrm{exp}},\\
D_{\mathrm{eff}}&=D\times L, & \Ecomp&=k\times D\times L,\\
\Eeq&=E\times D\times L, & |\mathcal{R}_{\mathrm{formal}}|&=\binom{E}{k}^{D\times L},
\end{align*}
where $\mathcal{R}_{\mathrm{formal}}$ is the set of formal routes of one token, the ordered sequences $(S_{t,d}^{(\ell)})_{d\leq D,\,\ell\leq L}$ of unordered top-$k$ selections; its size follows from the $\binom{E}{k}$ choices at each of the $D\times L$ visits. A trained model need not realise every route, and different routes can implement the same function; likewise, $\Eeq$ counts parameter-tied expert slots in the unrolled computation, not independently learned experts. Holding $E\times D$, $D\times L$ and $k$ fixed while increasing $E$ enlarges $|\mathcal{R}_{\mathrm{formal}}|$ without increasing the number of expert calls, but this alone establishes neither greater functional capacity nor lower loss.

\paragraph{Bounds on expert coverage.}
The number of distinct experts token $t$ reaches over its passes satisfies
\begin{equation}
k\times D\;\leq\;U_t=\sum_{d=1}^{D}\Bigl|\bigcup_{\ell=1}^{L}S_{t,d}^{(\ell)}\Bigr|\;\leq\;D\min(E,kL).
\label{eq:coverage-bounds}
\end{equation}
In each layer the union contains the $k$ distinct experts of any one visit, and it contains at most the $E$ experts of the layer and at most the $kL$ selections made over the $L$ visits; summing over the $D$ layers gives both bounds.

\paragraph{Three comparison regimes.}
The identities separate three ways of comparing looped MoE models.
\begin{itemize}
\item \emph{Additional passes at fixed parameters.} Holding $E$ and $D$ fixed while increasing $L$ preserves the stored parameters when all block parameters are shared and no pass-specific parameters are introduced. Effective depth and expert calls increase in proportion to $L$. This comparison measures the benefit of additional computation through parameter reuse.
\item \emph{Parameter--computation trade-offs.} Holding $E$ and $D\times L$ fixed while reducing $D$ and increasing $L$ preserves the number of layer evaluations and expert calls but reduces the number of stored experts. With unchanged sublayer dimensions and sequence length, the leading forward arithmetic is matched.
\item \emph{Flattening at fixed expert parameters.} For an integer $a\geq1$ dividing $D$, the map $(E,D,L)\mapsto(aE,D/a,aL)$ preserves $E\times D$, $D\times L$ and $k\times D\times L$ while multiplying $\Eeq$ by $a$. The per-layer cap on distinct experts rises from $\min(E,kL)$ to $a\min(E,kL)$, whereas the whole-block cap stays $D\min(E,kL)$. Flattening therefore enlarges the pool available at each routing decision without raising the maximum number of distinct experts a token can reach across the block. Each flattening step of \Cref{sec:method-foil} uses $a=2$.
\end{itemize}

\paragraph{Total parameters of Foil.}
The total parameter count of Foil is
\begin{equation}
P_{\mathrm{total}}
= P_{\mathrm{outside}}
+ E\times D\times p_{\mathrm{exp}}
+ D\times L\times P_{\mathrm{attn}}
+ D\times P_{\mathrm{router}}(E),
\label{eq:foil-params}
\end{equation}
where $P_{\mathrm{outside}}$ counts the parameters outside the looped block and the remaining terms count its experts, attention and routers (our RMSNorm layers have no gain and hence no parameters). With attention shared across passes (proto-Foil), the attention term is $D\times P_{\mathrm{attn}}$ instead. For a bias-free linear router of input width $\dmodel$, $P_{\mathrm{router}}(E)=\dmodel\times E$, so the router total is also preserved by flattening. Since $D\times L$ is fixed along the flattening sequence, Foil keeps the attention parameters, and hence the total, unchanged, whereas with shared attention they decrease with $D$. Likewise, fixed $D\times L$ and $k\times D\times L$ preserve the leading attention and selected-expert arithmetic, but not router scoring: a linear router produces $E\times D\times L$ expert scores per token, at a cost of $O(E\times D\times L\times\dmodel)$, which grows along the sequence. $\Ecomp$ therefore measures expert computation, not total FLOPs including routing and dispatch.

\section{Expert-collapse diagnostics: definitions and limits}
\label{sec:app-metrics}

All quantities are empirical summaries of the same $N=255{,}500$ probe tokens at the final checkpoint, restricted to the recurrent core. Index tokens by $t$, physical layers of the core by $d$ and passes by $\ell$, as in \Cref{sec:method}. For each decision, the finite router logits $z_{t,d,i}^{(\ell)}$ define $p_{t,d,i}^{(\ell)}=\exp(z_{t,d,i}^{(\ell)})/\sum_j\exp(z_{t,d,j}^{(\ell)})$. The recorded top-$k$ set $S_{t,d}^{(\ell)}$ contains exactly $k=2$ distinct experts, with ties resolved by the model's routing rule. Load uses these selections, not probability mass or the renormalised mixture weights.

\paragraph{Class 0: coverage of physical experts.}
The distinct-expert count is
\begin{equation}
U_t=\sum_{d=1}^{D}\left|\bigcup_{\ell=1}^{L}S_{t,d}^{(\ell)}\right|,
\qquad kD\leq U_t\leq D\min(E,kL).
\label{eq:expert-coverage}
\end{equation}
For the bounds see \Cref{eq:coverage-bounds}. We report $\bar U=N^{-1}\sum_t U_t$ and $\bar U/[D\min(E,kL)]$. At $L=1$ the ratio is identically one and is omitted from \Cref{tab:app-metrics}.
Along the flattening sequence the number of distinct experts a token reaches falls (Base $24.1$, Foil-3 $17.8$, Foil-2 $12.5$, Foil-1 $9.7$ at 20B; \Cref{tab:app-metrics}): flattening enlarges the pool of each routing decision, not the number of experts a token uses. The increase with additional passes reported in the introduction (\Cref{fig:motivation}) is at fixed shape.

\paragraph{Class 1: load concentration.}
The count share and effective expert count of layer $d$ are
\begin{equation}
q_{d i}=\frac{1}{NLk}\sum_{t=1}^{N}\sum_{\ell=1}^{L}
\mathbf{1}\{i\in S_{t,d}^{(\ell)}\},
\qquad N_{2,d}=\frac{1}{\sum_{i=1}^{E}q_{d i}^{2}},
\qquad B_{2,d}=\frac{N_{2,d}}{E}.
\label{eq:expert-load}
\end{equation}
$N_{2,d}$ is the inverse Simpson effective count~\citep[Eq.~S10]{wu2026reba}; it equals $s$ for traffic uniformly spread over $s$ experts. Its normalisation $B_{2,d}$ is Jain's fairness index~\citep{jain1984fairness}. Since $\sum_iq_{d i}=1$ and $0\leq q_{d i}\leq1/k$, Cauchy--Schwarz and $q_{d i}^2\leq q_{d i}/k$ give
\begin{equation}
\frac{1}{E}\leq\sum_iq_{d i}^{2}\leq\frac{1}{k},
\qquad \frac{k}{E}\leq B_{2,d}\leq1.
\label{eq:load-bounds}
\end{equation}
$B_{2,d}=1$ if and only if the load is uniform. The lower endpoint $B_{2,d}=k/E$ holds if and only if the same $k$ experts are selected at every decision in that layer, the maximally concentrated load permitted by top-$k$ dispatch. Intermediate values quantify concentration without specifying a universal failure threshold. Unlike MaxVio, which measures the largest relative overload~\citep[Eq.~4]{wang2024auxfree}, $B_2$ depends on all expert shares.

Counts are pooled over passes before taking the reciprocal. The model score is
\begin{equation}
B_{\mathrm{model},2}=\frac{\sum_{d=1}^{D}N_{2,d}}{DE}
=\frac{1}{D}\sum_{d=1}^{D}B_{2,d}.
\label{eq:model-load-balance}
\end{equation}
Pooling can conceal concentration within a pass: when $E=kL$, partition the experts into $L$ disjoint groups of $k$ and assign all tokens to group $\ell$ on pass $\ell$. This gives pooled $B_2=1$, although each pass has $B_2=k/E$. A per-pass score instead uses $q_{d,\ell,i}=(Nk)^{-1}\sum_t\mathbf{1}\{i\in S_{t,d}^{(\ell)}\}$; averaging these scores generally differs from pooling counts.

\paragraph{Class 2: median margin ratio.}
Suppress the decision indices and order logits as $z_{(1)}\geq\cdots\geq z_{(E)}$. With $m=\lceil E/2\rceil$, define
\begin{equation}
R=\frac{p_{(1)}}{p_{(m)}}=\exp\bigl(z_{(1)}-z_{(m)}\bigr),
\qquad
\operatorname{\TMCR}(\mathcal I)
=\exp\!\left(\frac{1}{|\mathcal I|}\sum_{a\in\mathcal I}
\bigl[z_{a,(1)}-z_{a,(m)}\bigr]\right).
\label{eq:tmcr}
\end{equation}
All experimental pools have even $E$: the denominator is the upper of the two central probabilities, at descending rank $E/2$, rather than their arithmetic mean. The nonempty index set $\mathcal I$ contains the token--layer--pass decisions being summarised; the full-core score weights all $NDL$ decisions equally. This is the geometric mean of decision-level ratios, not a ratio of averaged probabilities. The logit form cancels the softmax normaliser and avoids division by probabilities that may underflow.

\TMCR{} is at least one and equals one exactly when the top $m$ logits coincide at every included decision. Complete router indifference, $p_i=1/E$ for every expert at every decision, is sufficient but not necessary: $z=(0,0,0,0,-c,-c,-c,-c)$ with $c>0$ also gives $R=1$. Replacing every logit vector by $\alpha z+\beta\mathbf{1}$, with $\alpha>0$ and the same tie rule, leaves top-$k$ selections and $B_2$ unchanged but sends \TMCR{} to $\operatorname{\TMCR}^{\alpha}$. \TMCR{} measures separation from the median rank and depends on logit scale; it does not measure the margin between the selected and unselected experts.

\paragraph{Joint interpretation.}
Load concentration and weak router preference are distinct routing phenomena. They also differ from the collapse of hidden representations studied by \citet{chi2022representation}. Neither $B_2$ nor \TMCR{} observes expert outputs: even identical expert functions can coexist with balanced, confident routing. Consequently, the scores describe traffic and router preference, while the masking test in Appendix~\ref{sec:app-masking} measures sensitivity to expert removal. \Cref{tab:app-metrics} reports the diagnostics; comparisons use the same probe and aggregation, and absolute comparisons in the main text are restricted to equal shapes because both scores depend on $E$.

\paragraph{Related metrics used in prior work.}
Besides the three metrics of \Cref{sec:method-metrics}, the literature uses further load-balance and routing-confidence metrics; \Cref{tab:app-related-metrics} lists the ones we also computed, with their ranges under top-$k$ routing. For one routing decision, $p_i$ are the router's softmax probabilities over the $E$ experts (temperature $1$), sorted as $p_{(1)}\geq\cdots\geq p_{(E)}$; $S$ is the selected set and $a_i=p_i/\sum_{j\in S}p_j$ the renormalised weight of $i\in S$; $q_i$ are the load shares of \Cref{eq:expert-load}. Our models use no capacity limit, so no assignment is ever dropped; \Cref{tab:app-related-values} gives the values of the other related metrics for every small-tier model.

\begin{table}[h]
\centering
\caption{Related metrics used in prior work: load balance (top four rows) and routing confidence (bottom five rows). Ranges are for top-$k$ routing with $k<E$. Cited works show prior use of each quantity; the exact formulas and normalisations are as stated here.}
\label{tab:app-related-metrics}
\resizebox{\linewidth}{!}{%
\begin{tabular}{llll}
\toprule
Metric & Definition & Range, direction & Source \\
\midrule
normalised load entropy $B_H$ & $-\sum_i q_i\ln q_i/\ln E$ & $[\ln k/\ln E,1]$, higher more even & \citep{nguyen2026libmoe} \\
MaxVio & $E\max_i q_i-1$ & $[0,E/k-1]$, lower less overload & \citep{wang2024auxfree} \\
normalised Gini $G^\ast$ & $\sum_{i,j}|q_i-q_j|/[2(E-1)]$ & $[0,(E-k)/(E-1)]$, lower more even & \citep{chen2026phibalancing} \\
dropped-assignment rate & dropped / requested assignments & $[0,1]$; not applicable (no drops) & \citep{fedus2022switch,lepikhin2021gshard} \\
\midrule
selected-expert probability & $p_i$, $i\in S$ & $[0,1]$, higher more support & \citep{fedus2022switch,riquelme2021vmoe} \\
selected-set mass & $\sum_{i\in S}p_i$ & $[k/E,1]$, higher more concentrated & \citep{shahout2025laser} \\
boundary margin & $\ln p_{(k)}-\ln p_{(k+1)}$ & $[0,\infty)$, higher clearer selection & \citep{thaman2025experts} \\
full-pool entropy concentration & $1-H(p)/\ln E$ & $[0,1]$, higher sharper & \citep{shahout2025laser} \\
selected-weight concentration & $1-H(a)/\ln k$ & $[0,1]$, higher one expert dominates & \citep{nguyen2026libmoe} \\
\bottomrule
\end{tabular}}
\end{table}

\section{Experimental configuration}  %
\label{sec:app-training}

\paragraph{Data.}
All models are trained on the \texttt{sample-100BT} subset of FineWeb-Edu~\citep{penedo2024fineweb}, tokenised with the SmolLM2 tokenizer~\citep{allal2025smollm2} (vocabulary $49{,}152$), which yields $101.7$B tokens. Sequences are packed to a context length of $4{,}096$ tokens. Every run reads the data in the same fixed order, so that at every optimiser step all runs have seen the same tokens; this makes the step-wise paired comparisons possible.

\paragraph{Optimisation.}
We use AdamW~\citep{loshchilov2019adamw} with $\beta_1=0.9$, $\beta_2=0.95$, weight decay $0.1$ and gradient clipping at norm $1.0$, in bfloat16. The global batch is $96$ sequences, i.e.\ $393{,}216$ tokens per step. The learning rate follows a warmup--stable--decay schedule~\citep{hu2024minicpm}: a linear warmup over $1{,}000$ steps to a peak of $3\times10^{-4}$, a constant phase, and a decay over the last $10\%$ of the run with a $1-\sqrt{\cdot}$ shape down to $5\%$ of the peak. All models share this schedule, the initialisation seed ($42$) and the data order; the seed replicate changes only the initialisation seed, to $43$.

\paragraph{Auxiliary losses.}
Every model adds to the language-modelling loss a load-balancing loss of the Switch Transformer form~\citep{fedus2022switch} with coefficient $0.01$ and a router z-loss~\citep{zoph2022stmoe} with coefficient $0.001$. Both are computed for every MoE layer on every pass and averaged over the unrolled depth. The losses we report are the language-modelling loss alone.

\paragraph{Training length.}
The 20B-token stage runs $50{,}000$ steps ($19.66$B tokens); the constant phase ends at step $45{,}000$ and the decay occupies the remaining $5{,}000$ steps. The 100B-token continuation starts from the step-$45{,}000$ checkpoint, taken before the decay, resumes the optimiser state and the data order, and lengthens the schedule to $254{,}313$ steps ($100.0$B tokens), with the learning rate at its peak until step $228{,}882$ and the same decay afterwards.

\paragraph{Reported loss.}
The reported loss of a model is its mean language-modelling loss over the last $2{,}000$ training steps, and two models are compared by the mean of their step-wise loss difference over this window. Loss standard errors treat consecutive training steps as independent and are therefore understated.

\begin{table}[t]
\centering
\footnotesize
\setlength{\tabcolsep}{3pt}
\caption{Per-configuration hyperparameters and parameter counts. $(E,D,L)$: experts per MoE layer, layers in the loop block, loop passes. $E_{\text{real}} = E\cdot D$ (distinct experts), $E_{\text{comp}} = D\cdot L\cdot k$ (expert calls per token), $E_{\text{eq}} = E\cdot D\cdot L$. Attention: \emph{shared} = one set of attention weights reused by every pass; \emph{untied} = one set per pass. Parameter columns in millions, rounded independently: total; experts (including their routers) and attention inside the loop block; the non-looped head and tail layers; ``other'' = token embeddings and output projection. Run ids are the identifiers used in our logs. Base width: $d_{\text{model}}=1024$, 16 attention heads, expert hidden width 1536, top-$k=2$, one head and one tail layer with 8 experts each, vocabulary 49{,}152.
$^{a}$\,$E=4$ with $k=2$: half of the experts in a layer are active for every token. $^{b}$\,Two head and two tail layers. $^{c}$\,Not looped ($L=1$).}
\label{tab:app-hparams}
\begin{tabular}{@{}l l c c r r r r r r r r@{}}
\toprule
Model & Run id & $(E,D,L)$ & Attention & $E_{\text{real}}$ & $E_{\text{comp}}$ & $E_{\text{eq}}$ & Total & \begin{tabular}[b]{@{}c@{}}Loop\\experts\end{tabular} & \begin{tabular}[b]{@{}c@{}}Loop\\attn.\end{tabular} & \begin{tabular}[b]{@{}c@{}}Head/\\tail\end{tabular} & Other \\
\midrule
\multicolumn{12}{@{}l}{\emph{Base width ($d_{\text{model}}=1024$): Foil family}} \\
Base-tied & S1 & (8,8,2) & shared & 64 & 32 & 128 & 520.2 & 302.1 & 33.6 & 83.9 & 100.7 \\
proto-Foil-3 & S2 & (16,4,4) & shared & 64 & 32 & 256 & 503.4 & 302.1 & 16.8 & 83.9 & 100.7 \\
proto-Foil-2 & S3 & (32,2,8) & shared & 64 & 32 & 512 & 495.0 & 302.1 & 8.4 & 83.9 & 100.7 \\
proto-Foil-1 & S4 & (64,1,16) & shared & 64 & 32 & 1024 & 490.8 & 302.1 & 4.2 & 83.9 & 100.7 \\
Base & U1 & (8,8,2) & untied & 64 & 32 & 128 & 553.7 & 302.1 & 67.1 & 83.9 & 100.7 \\
Foil-3 & U2 & (16,4,4) & untied & 64 & 32 & 256 & 553.7 & 302.1 & 67.1 & 83.9 & 100.7 \\
Foil-2 & U3 & (32,2,8) & untied & 64 & 32 & 512 & 553.7 & 302.1 & 67.1 & 83.9 & 100.7 \\
Foil-1 & U4 & (64,1,16) & untied & 64 & 32 & 1024 & 553.7 & 302.1 & 67.1 & 83.9 & 100.7 \\
\addlinespace
\multicolumn{12}{@{}l}{\emph{Base width ($d_{\text{model}}=1024$): other grid models}} \\
-- & S6$^{c}$ & (8,8,1) & shared & 64 & 16 & 64 & 520.2 & 302.1 & 33.6 & 83.9 & 100.7 \\
-- & S8 & (8,8,3) & shared & 64 & 48 & 192 & 520.2 & 302.1 & 33.6 & 83.9 & 100.7 \\
-- & S9 & (8,8,4) & shared & 64 & 64 & 256 & 520.2 & 302.1 & 33.6 & 83.9 & 100.7 \\
-- & N4 & (8,8,8) & shared & 64 & 128 & 512 & 520.2 & 302.1 & 33.6 & 83.9 & 100.7 \\
-- & S5 & (16,4,2) & shared & 64 & 16 & 128 & 503.4 & 302.1 & 16.8 & 83.9 & 100.7 \\
-- & N3 & (16,4,6) & shared & 64 & 48 & 384 & 503.4 & 302.1 & 16.8 & 83.9 & 100.7 \\
-- & N2 & (16,4,8) & shared & 64 & 64 & 512 & 503.4 & 302.1 & 16.8 & 83.9 & 100.7 \\
-- & S7 & (8,4,2) & shared & 32 & 16 & 64 & 352.4 & 151.0 & 16.8 & 83.9 & 100.7 \\
-- & S12 & (8,4,4) & shared & 32 & 32 & 128 & 352.4 & 151.0 & 16.8 & 83.9 & 100.7 \\
-- & N1 & (8,4,8) & shared & 32 & 64 & 256 & 352.4 & 151.0 & 16.8 & 83.9 & 100.7 \\
-- & S11$^{a}$ & (4,8,2) & shared & 32 & 32 & 64 & 369.1 & 151.0 & 33.6 & 83.9 & 100.7 \\
-- & S13$^{a}$ & (4,8,4) & shared & 32 & 64 & 128 & 369.1 & 151.0 & 33.6 & 83.9 & 100.7 \\
-- & S10 & (16,8,2) & shared & 128 & 32 & 256 & 822.2 & 604.1 & 33.6 & 83.9 & 100.7 \\
-- & S14$^{c}$ & (8,16,1) & shared & 128 & 32 & 128 & 855.8 & 604.1 & 67.1 & 83.9 & 100.7 \\
-- & S15$^{b}$ & (8,8,2) & shared & 64 & 32 & 128 & 604.1 & 302.1 & 33.6 & 167.8 & 100.7 \\
\bottomrule
\end{tabular}
\end{table}

\section{Experimental results}  %
\label{sec:app-full}

\paragraph{Seed replicate.}
\label{sec:app-seed}
Base-tied was retrained with only the initialisation seed changed (seed 43 instead of 42; both runs are in \Cref{tab:app-main}). The paired difference of the final-window loss is $-0.0003\pm0.0009$~nat, and every difference we interpret as a result is at least $0.002$~nat; smaller differences are reported as on par.

\Cref{tab:app-main} collects the configuration, parameter count and final-window loss of every small-tier model. The three expert counts are $\Ereal=ED$ (distinct experts in the core), $\Ecomp=DLk$ (expert calls per token) and $\Eeq=EDL$ (experts of the equivalent non-looped model).

\begin{table}[h]
\centering
\caption{Main results of all small-tier models ($\dmodel=1024$) at 20B tokens, and at 100B tokens where the model was continued. Loss: mean task loss over the final window (nat); perplexity is its exponential. Parameters in millions. $^\S$Repeated from an earlier group for comparison. Single seed except the seed-43 repeat of Base-tied.}
\label{tab:app-main}
\resizebox{\linewidth}{!}{%
\begin{tabular}{lccccccccc}
\toprule
Model & $(E,D,L)$ & attention & params (M) & $\Ereal$ & $\Ecomp$ & $\Eeq$ & loss 20B & ppl 20B & loss 100B \\
\midrule
\multicolumn{10}{l}{\emph{Flattening, tied attention}} \\
Base-tied & $(8,8,2)$ & tied & $520.2$ & 64 & 32 & 128 & $2.688$ & $14.71$ & $2.522$ \\
proto-Foil-3 & $(16,4,4)$ & tied & $503.4$ & 64 & 32 & 256 & $2.679$ & $14.58$ & $2.520$ \\
proto-Foil-2 & $(32,2,8)$ & tied & $495.0$ & 64 & 32 & 512 & $2.684$ & $14.64$ & $2.531$ \\
proto-Foil-1 & $(64,1,16)$ & tied & $490.8$ & 64 & 32 & 1024 & $2.700$ & $14.89$ & $2.548$ \\
\midrule
\multicolumn{10}{l}{\emph{Flattening, untied attention}} \\
Base & $(8,8,2)$ & untied & $553.7$ & 64 & 32 & 128 & $2.666$ & $14.38$ & $2.511$ \\
Foil-3 & $(16,4,4)$ & untied & $553.7$ & 64 & 32 & 256 & $2.663$ & $14.33$ & $2.506$ \\
Foil-2 & $(32,2,8)$ & untied & $553.7$ & 64 & 32 & 512 & $2.657$ & $14.25$ & $2.501$ \\
Foil-1 & $(64,1,16)$ & untied & $553.7$ & 64 & 32 & 1024 & $2.659$ & $14.28$ & $2.499$ \\
\midrule
\multicolumn{10}{l}{\emph{Loop passes, $(E,D)=(8,8)$}} \\
-- & $(8,8,1)$ & tied & $520.2$ & 64 & 16 & 64 & $2.752$ & $15.68$ & $2.582$ \\
Base-tied$^\S$ & $(8,8,2)$ & tied & $520.2$ & 64 & 32 & 128 & $2.688$ & $14.71$ & $2.522$ \\
-- & $(8,8,3)$ & tied & $520.2$ & 64 & 48 & 192 & $2.650$ & $14.16$ & -- \\
-- & $(8,8,4)$ & tied & $520.2$ & 64 & 64 & 256 & $2.629$ & $13.86$ & -- \\
-- & $(8,8,8)$ & tied & $520.2$ & 64 & 128 & 512 & $2.596$ & $13.41$ & -- \\
\midrule
\multicolumn{10}{l}{\emph{Loop passes, $(E,D)=(16,4)$}} \\
-- & $(16,4,2)$ & tied & $503.4$ & 64 & 16 & 128 & $2.749$ & $15.63$ & -- \\
proto-Foil-3$^\S$ & $(16,4,4)$ & tied & $503.4$ & 64 & 32 & 256 & $2.679$ & $14.58$ & $2.520$ \\
-- & $(16,4,6)$ & tied & $503.4$ & 64 & 48 & 384 & $2.649$ & $14.14$ & -- \\
-- & $(16,4,8)$ & tied & $503.4$ & 64 & 64 & 512 & $2.638$ & $13.99$ & -- \\
\midrule
\multicolumn{10}{l}{\emph{Loop passes, $(E,D)=(8,4)$}} \\
-- & $(8,4,2)$ & tied & $352.4$ & 32 & 16 & 64 & $2.787$ & $16.24$ & -- \\
-- & $(8,4,4)$ & tied & $352.4$ & 32 & 32 & 128 & $2.720$ & $15.17$ & -- \\
-- & $(8,4,8)$ & tied & $352.4$ & 32 & 64 & 256 & $2.688$ & $14.70$ & -- \\
\midrule
\multicolumn{10}{l}{\emph{Pool size and depth at fixed $\Ecomp=32$}} \\
-- & $(4,8,2)$ & tied & $369.1$ & 32 & 32 & 64 & $2.728$ & $15.30$ & -- \\
Base-tied$^\S$ & $(8,8,2)$ & tied & $520.2$ & 64 & 32 & 128 & $2.688$ & $14.71$ & $2.522$ \\
-- & $(16,8,2)$ & tied & $822.2$ & 128 & 32 & 256 & $2.641$ & $14.03$ & -- \\
--$^\S$ & $(8,4,4)$ & tied & $352.4$ & 32 & 32 & 128 & $2.720$ & $15.17$ & -- \\
-- & $(8,16,1)$ & tied & $855.8$ & 128 & 32 & 128 & $2.625$ & $13.81$ & $2.463$ \\
\midrule
\multicolumn{10}{l}{\emph{Other controls}} \\
-- & $(4,8,4)$ & tied & $369.1$ & 32 & 64 & 128 & $2.675$ & $14.51$ & -- \\
Base-tied, two-layer prelude/coda & $(8,8,2)$ & tied & $604.1$ & 64 & 32 & 128 & $2.656$ & $14.24$ & -- \\
Base-tied, seed 43 & $(8,8,2)$ & tied & $520.2$ & 64 & 32 & 128 & $2.688$ & $14.70$ & -- \\
\bottomrule
\end{tabular}}
\end{table}

\section{Downstream evaluation}
\label{sec:app-eval}

Final checkpoints are evaluated on 36 downstream tasks. A task is kept as valid only if all 35 models trained for 20B tokens (all widths) score, on average, at least five standard errors above its baseline (chance level, zero for open-ended completion, or the majority class for classification tasks), in the metric and setting that lies furthest above the baseline. The appendix reports seven valid tasks: the three representative tasks of the main text, and PROST, LAMBADA (OpenAI), SWAG and BLiMP.
The three representative tasks of the main text take one task from each of three categories: word prediction (LAMBADA, standard split), sentence continuation (HellaSwag) and coreference resolution (XWinograd, English). Zero-shot is the main setting (\Cref{tab:app-eval-0shot}). For 5-shot (\Cref{tab:app-eval-5shot}), HellaSwag is averaged over three evaluation seeds, which change the choice of in-context examples; the other tasks have seed~0 only, and BLiMP and LAMBADA (OpenAI) have no 5-shot variant. Standard errors are those of the evaluation harness; a mean over tasks treats the tasks as independent, and the standard error of a difference between two models is $\sqrt{\mathrm{se}_a^2+\mathrm{se}_b^2}$, which ignores that both models answer the same questions and is therefore conservative.

\begin{table}[h]
\centering
\caption{Zero-shot downstream accuracy (\%, $\pm1$ standard error) of the eight models of \Cref{tab:app-hparams}. SWAG, HellaSwag and PROST are scored by length-normalised accuracy, the other tasks by accuracy; the mean over tasks treats tasks as independent (standard error of a difference between two models about $0.31$).}
\label{tab:app-eval-0shot}
\resizebox{\linewidth}{!}{%
\begin{tabular}{lcccccccc}
\toprule
Task & Base-tied & proto-Foil-3 & proto-Foil-2 & proto-Foil-1 & Base & Foil-3 & Foil-2 & Foil-1 \\
\midrule
\multicolumn{9}{l}{\emph{20B tokens}} \\
PROST & $31.27$\,{\scriptsize$\pm0.34$} & $31.00$\,{\scriptsize$\pm0.34$} & $29.50$\,{\scriptsize$\pm0.33$} & $30.73$\,{\scriptsize$\pm0.34$} & $30.26$\,{\scriptsize$\pm0.34$} & $31.71$\,{\scriptsize$\pm0.34$} & $31.10$\,{\scriptsize$\pm0.34$} & $29.80$\,{\scriptsize$\pm0.33$} \\
LAMBADA (standard) & $23.17$\,{\scriptsize$\pm0.59$} & $24.22$\,{\scriptsize$\pm0.60$} & $25.48$\,{\scriptsize$\pm0.61$} & $23.73$\,{\scriptsize$\pm0.59$} & $25.89$\,{\scriptsize$\pm0.61$} & $26.14$\,{\scriptsize$\pm0.61$} & $26.66$\,{\scriptsize$\pm0.62$} & $26.14$\,{\scriptsize$\pm0.61$} \\
LAMBADA (OpenAI) & $32.08$\,{\scriptsize$\pm0.65$} & $32.33$\,{\scriptsize$\pm0.65$} & $32.23$\,{\scriptsize$\pm0.65$} & $31.09$\,{\scriptsize$\pm0.64$} & $35.30$\,{\scriptsize$\pm0.67$} & $34.54$\,{\scriptsize$\pm0.66$} & $33.86$\,{\scriptsize$\pm0.66$} & $34.29$\,{\scriptsize$\pm0.66$} \\
SWAG & $55.34$\,{\scriptsize$\pm0.35$} & $55.47$\,{\scriptsize$\pm0.35$} & $55.36$\,{\scriptsize$\pm0.35$} & $54.69$\,{\scriptsize$\pm0.35$} & $55.74$\,{\scriptsize$\pm0.35$} & $55.90$\,{\scriptsize$\pm0.35$} & $56.73$\,{\scriptsize$\pm0.35$} & $56.41$\,{\scriptsize$\pm0.35$} \\
HellaSwag & $39.55$\,{\scriptsize$\pm0.49$} & $40.41$\,{\scriptsize$\pm0.49$} & $40.72$\,{\scriptsize$\pm0.49$} & $39.86$\,{\scriptsize$\pm0.49$} & $40.82$\,{\scriptsize$\pm0.49$} & $41.30$\,{\scriptsize$\pm0.49$} & $41.07$\,{\scriptsize$\pm0.49$} & $41.46$\,{\scriptsize$\pm0.49$} \\
BLiMP & $82.18$\,{\scriptsize$\pm0.13$} & $82.07$\,{\scriptsize$\pm0.13$} & $81.09$\,{\scriptsize$\pm0.14$} & $81.15$\,{\scriptsize$\pm0.14$} & $81.94$\,{\scriptsize$\pm0.13$} & $81.98$\,{\scriptsize$\pm0.13$} & $82.14$\,{\scriptsize$\pm0.13$} & $82.40$\,{\scriptsize$\pm0.13$} \\
XWinograd (en) & $63.78$\,{\scriptsize$\pm1.00$} & $63.01$\,{\scriptsize$\pm1.00$} & $63.61$\,{\scriptsize$\pm1.00$} & $62.37$\,{\scriptsize$\pm1.00$} & $64.52$\,{\scriptsize$\pm0.99$} & $64.09$\,{\scriptsize$\pm1.00$} & $64.77$\,{\scriptsize$\pm0.99$} & $64.56$\,{\scriptsize$\pm0.99$} \\
\midrule
Mean, seven tasks & $46.77$\,{\scriptsize$\pm0.22$} & $46.93$\,{\scriptsize$\pm0.22$} & $46.86$\,{\scriptsize$\pm0.22$} & $46.23$\,{\scriptsize$\pm0.21$} & $47.78$\,{\scriptsize$\pm0.22$} & $47.95$\,{\scriptsize$\pm0.22$} & $48.05$\,{\scriptsize$\pm0.22$} & $47.87$\,{\scriptsize$\pm0.22$} \\
Mean, three representative tasks & $42.17$\,{\scriptsize$\pm0.42$} & $42.55$\,{\scriptsize$\pm0.42$} & $43.27$\,{\scriptsize$\pm0.42$} & $41.99$\,{\scriptsize$\pm0.42$} & $43.74$\,{\scriptsize$\pm0.42$} & $43.84$\,{\scriptsize$\pm0.42$} & $44.17$\,{\scriptsize$\pm0.42$} & $44.05$\,{\scriptsize$\pm0.42$} \\
\midrule
\multicolumn{9}{l}{\emph{100B tokens}} \\
PROST & $29.84$\,{\scriptsize$\pm0.33$} & $31.21$\,{\scriptsize$\pm0.34$} & $31.85$\,{\scriptsize$\pm0.34$} & $30.34$\,{\scriptsize$\pm0.34$} & $29.60$\,{\scriptsize$\pm0.33$} & $34.52$\,{\scriptsize$\pm0.35$} & $29.92$\,{\scriptsize$\pm0.33$} & $29.45$\,{\scriptsize$\pm0.33$} \\
LAMBADA (standard) & $29.85$\,{\scriptsize$\pm0.64$} & $29.32$\,{\scriptsize$\pm0.63$} & $28.31$\,{\scriptsize$\pm0.63$} & $26.57$\,{\scriptsize$\pm0.62$} & $30.43$\,{\scriptsize$\pm0.64$} & $31.38$\,{\scriptsize$\pm0.65$} & $30.18$\,{\scriptsize$\pm0.64$} & $31.83$\,{\scriptsize$\pm0.65$} \\
LAMBADA (OpenAI) & $36.44$\,{\scriptsize$\pm0.67$} & $38.42$\,{\scriptsize$\pm0.68$} & $37.16$\,{\scriptsize$\pm0.67$} & $36.23$\,{\scriptsize$\pm0.67$} & $39.24$\,{\scriptsize$\pm0.68$} & $39.80$\,{\scriptsize$\pm0.68$} & $39.61$\,{\scriptsize$\pm0.68$} & $39.98$\,{\scriptsize$\pm0.68$} \\
SWAG & $59.13$\,{\scriptsize$\pm0.35$} & $59.32$\,{\scriptsize$\pm0.35$} & $58.77$\,{\scriptsize$\pm0.35$} & $58.44$\,{\scriptsize$\pm0.35$} & $59.17$\,{\scriptsize$\pm0.35$} & $59.98$\,{\scriptsize$\pm0.35$} & $59.90$\,{\scriptsize$\pm0.35$} & $59.98$\,{\scriptsize$\pm0.35$} \\
HellaSwag & $47.17$\,{\scriptsize$\pm0.50$} & $47.07$\,{\scriptsize$\pm0.50$} & $46.59$\,{\scriptsize$\pm0.50$} & $45.68$\,{\scriptsize$\pm0.50$} & $47.23$\,{\scriptsize$\pm0.50$} & $47.41$\,{\scriptsize$\pm0.50$} & $48.06$\,{\scriptsize$\pm0.50$} & $48.43$\,{\scriptsize$\pm0.50$} \\
BLiMP & $82.12$\,{\scriptsize$\pm0.13$} & $81.14$\,{\scriptsize$\pm0.14$} & $80.24$\,{\scriptsize$\pm0.14$} & $81.64$\,{\scriptsize$\pm0.13$} & $81.97$\,{\scriptsize$\pm0.14$} & $82.13$\,{\scriptsize$\pm0.13$} & $81.66$\,{\scriptsize$\pm0.13$} & $81.82$\,{\scriptsize$\pm0.13$} \\
XWinograd (en) & $67.87$\,{\scriptsize$\pm0.97$} & $68.26$\,{\scriptsize$\pm0.97$} & $66.92$\,{\scriptsize$\pm0.98$} & $67.01$\,{\scriptsize$\pm0.98$} & $68.22$\,{\scriptsize$\pm0.97$} & $68.95$\,{\scriptsize$\pm0.96$} & $70.67$\,{\scriptsize$\pm0.94$} & $68.82$\,{\scriptsize$\pm0.96$} \\
\midrule
Mean, seven tasks & $50.35$\,{\scriptsize$\pm0.22$} & $50.68$\,{\scriptsize$\pm0.22$} & $49.98$\,{\scriptsize$\pm0.22$} & $49.42$\,{\scriptsize$\pm0.22$} & $50.84$\,{\scriptsize$\pm0.22$} & $52.02$\,{\scriptsize$\pm0.22$} & $51.43$\,{\scriptsize$\pm0.21$} & $51.47$\,{\scriptsize$\pm0.22$} \\
Mean, three representative tasks & $48.30$\,{\scriptsize$\pm0.42$} & $48.22$\,{\scriptsize$\pm0.42$} & $47.27$\,{\scriptsize$\pm0.42$} & $46.42$\,{\scriptsize$\pm0.42$} & $48.63$\,{\scriptsize$\pm0.42$} & $49.25$\,{\scriptsize$\pm0.42$} & $49.64$\,{\scriptsize$\pm0.41$} & $49.69$\,{\scriptsize$\pm0.42$} \\
\bottomrule
\end{tabular}}
\end{table}

\begin{table}[h]
\centering
\caption{5-shot downstream accuracy (\%, $\pm1$ standard error). HellaSwag is the mean over three evaluation seeds; $^\dagger$seed~0 only; -- marks tasks without a 5-shot variant. The mean over five tasks excludes these two and treats tasks as independent (standard error of a difference between two models about $0.38$). Metrics as in \Cref{tab:app-eval-0shot}.}
\label{tab:app-eval-5shot}
\resizebox{\linewidth}{!}{%
\begin{tabular}{lcccccccc}
\toprule
Task & Base-tied & proto-Foil-3 & proto-Foil-2 & proto-Foil-1 & Base & Foil-3 & Foil-2 & Foil-1 \\
\midrule
\multicolumn{9}{l}{\emph{20B tokens}} \\
PROST$^\dagger$ & $33.24$\,{\scriptsize$\pm0.34$} & $31.50$\,{\scriptsize$\pm0.34$} & $26.96$\,{\scriptsize$\pm0.32$} & $27.88$\,{\scriptsize$\pm0.33$} & $30.35$\,{\scriptsize$\pm0.34$} & $29.96$\,{\scriptsize$\pm0.33$} & $29.12$\,{\scriptsize$\pm0.33$} & $29.50$\,{\scriptsize$\pm0.33$} \\
LAMBADA (standard)$^\dagger$ & $22.71$\,{\scriptsize$\pm0.58$} & $23.17$\,{\scriptsize$\pm0.59$} & $24.84$\,{\scriptsize$\pm0.60$} & $22.34$\,{\scriptsize$\pm0.58$} & $24.86$\,{\scriptsize$\pm0.60$} & $24.41$\,{\scriptsize$\pm0.60$} & $24.80$\,{\scriptsize$\pm0.60$} & $24.24$\,{\scriptsize$\pm0.60$} \\
LAMBADA (OpenAI) & -- & -- & -- & -- & -- & -- & -- & -- \\
SWAG$^\dagger$ & $54.41$\,{\scriptsize$\pm0.35$} & $55.22$\,{\scriptsize$\pm0.35$} & $54.85$\,{\scriptsize$\pm0.35$} & $54.33$\,{\scriptsize$\pm0.35$} & $54.96$\,{\scriptsize$\pm0.35$} & $55.30$\,{\scriptsize$\pm0.35$} & $55.59$\,{\scriptsize$\pm0.35$} & $55.55$\,{\scriptsize$\pm0.35$} \\
HellaSwag & $40.13$\,{\scriptsize$\pm0.49$} & $41.12$\,{\scriptsize$\pm0.49$} & $41.01$\,{\scriptsize$\pm0.49$} & $40.17$\,{\scriptsize$\pm0.49$} & $40.79$\,{\scriptsize$\pm0.49$} & $41.15$\,{\scriptsize$\pm0.49$} & $41.50$\,{\scriptsize$\pm0.49$} & $41.38$\,{\scriptsize$\pm0.49$} \\
BLiMP & -- & -- & -- & -- & -- & -- & -- & -- \\
XWinograd (en)$^\dagger$ & $63.78$\,{\scriptsize$\pm1.00$} & $65.20$\,{\scriptsize$\pm0.99$} & $65.03$\,{\scriptsize$\pm0.99$} & $63.57$\,{\scriptsize$\pm1.00$} & $66.80$\,{\scriptsize$\pm0.98$} & $65.08$\,{\scriptsize$\pm0.99$} & $64.43$\,{\scriptsize$\pm0.99$} & $67.18$\,{\scriptsize$\pm0.97$} \\
\midrule
Mean, five tasks & $42.85$\,{\scriptsize$\pm0.27$} & $43.24$\,{\scriptsize$\pm0.27$} & $42.54$\,{\scriptsize$\pm0.27$} & $41.66$\,{\scriptsize$\pm0.27$} & $43.55$\,{\scriptsize$\pm0.27$} & $43.18$\,{\scriptsize$\pm0.27$} & $43.09$\,{\scriptsize$\pm0.27$} & $43.57$\,{\scriptsize$\pm0.27$} \\
Mean, three representative tasks & $42.21$\,{\scriptsize$\pm0.42$} & $43.16$\,{\scriptsize$\pm0.42$} & $43.63$\,{\scriptsize$\pm0.42$} & $42.03$\,{\scriptsize$\pm0.42$} & $44.15$\,{\scriptsize$\pm0.42$} & $43.55$\,{\scriptsize$\pm0.42$} & $43.58$\,{\scriptsize$\pm0.42$} & $44.27$\,{\scriptsize$\pm0.41$} \\
\midrule
\multicolumn{9}{l}{\emph{100B tokens}} \\
PROST$^\dagger$ & $27.84$\,{\scriptsize$\pm0.33$} & $27.87$\,{\scriptsize$\pm0.33$} & $29.36$\,{\scriptsize$\pm0.33$} & $25.24$\,{\scriptsize$\pm0.32$} & $29.58$\,{\scriptsize$\pm0.33$} & $31.61$\,{\scriptsize$\pm0.34$} & $28.49$\,{\scriptsize$\pm0.33$} & $28.25$\,{\scriptsize$\pm0.33$} \\
LAMBADA (standard)$^\dagger$ & $28.66$\,{\scriptsize$\pm0.63$} & $26.49$\,{\scriptsize$\pm0.61$} & $27.44$\,{\scriptsize$\pm0.62$} & $25.25$\,{\scriptsize$\pm0.61$} & $30.02$\,{\scriptsize$\pm0.64$} & $27.87$\,{\scriptsize$\pm0.62$} & $30.93$\,{\scriptsize$\pm0.64$} & $30.91$\,{\scriptsize$\pm0.64$} \\
LAMBADA (OpenAI) & -- & -- & -- & -- & -- & -- & -- & -- \\
SWAG$^\dagger$ & $58.83$\,{\scriptsize$\pm0.35$} & $59.02$\,{\scriptsize$\pm0.35$} & $58.60$\,{\scriptsize$\pm0.35$} & $58.13$\,{\scriptsize$\pm0.35$} & $58.48$\,{\scriptsize$\pm0.35$} & $58.96$\,{\scriptsize$\pm0.35$} & $59.36$\,{\scriptsize$\pm0.35$} & $59.33$\,{\scriptsize$\pm0.35$} \\
HellaSwag & $47.52$\,{\scriptsize$\pm0.50$} & $47.56$\,{\scriptsize$\pm0.50$} & $47.07$\,{\scriptsize$\pm0.50$} & $46.20$\,{\scriptsize$\pm0.50$} & $47.61$\,{\scriptsize$\pm0.50$} & $47.95$\,{\scriptsize$\pm0.50$} & $48.78$\,{\scriptsize$\pm0.50$} & $48.48$\,{\scriptsize$\pm0.50$} \\
BLiMP & -- & -- & -- & -- & -- & -- & -- & -- \\
XWinograd (en)$^\dagger$ & $69.81$\,{\scriptsize$\pm0.95$} & $71.14$\,{\scriptsize$\pm0.94$} & $69.68$\,{\scriptsize$\pm0.95$} & $68.09$\,{\scriptsize$\pm0.97$} & $71.53$\,{\scriptsize$\pm0.94$} & $71.31$\,{\scriptsize$\pm0.94$} & $72.77$\,{\scriptsize$\pm0.92$} & $72.13$\,{\scriptsize$\pm0.93$} \\
\midrule
Mean, five tasks & $46.53$\,{\scriptsize$\pm0.27$} & $46.42$\,{\scriptsize$\pm0.26$} & $46.43$\,{\scriptsize$\pm0.27$} & $44.58$\,{\scriptsize$\pm0.27$} & $47.44$\,{\scriptsize$\pm0.27$} & $47.54$\,{\scriptsize$\pm0.27$} & $48.07$\,{\scriptsize$\pm0.26$} & $47.82$\,{\scriptsize$\pm0.27$} \\
Mean, three representative tasks & $48.66$\,{\scriptsize$\pm0.41$} & $48.40$\,{\scriptsize$\pm0.41$} & $48.06$\,{\scriptsize$\pm0.41$} & $46.51$\,{\scriptsize$\pm0.42$} & $49.72$\,{\scriptsize$\pm0.41$} & $49.04$\,{\scriptsize$\pm0.41$} & $50.83$\,{\scriptsize$\pm0.41$} & $50.51$\,{\scriptsize$\pm0.41$} \\
\bottomrule
\end{tabular}}
\end{table}

\section{Held-out validation loss}
\label{sec:app-heldout}
Held-out losses are computed at the final checkpoints on a FineWeb-Edu validation slice of $8.0$M tokens, disjoint from the training data, and on Wikipedia, C4 and arXiv slices of $3$--$4$M tokens each; the evaluation reports means only (\Cref{tab:app-heldout}).

\begin{table}[t]
\centering
\small
\caption{Held-out loss (nat) of the Foil family on a FineWeb-Edu validation slice and three domain slices, final checkpoints. The FineWeb-Edu slice reproduces the training-stream ordering: Foil-2 lowest at 20B, Foil-1 lowest at 100B, and the untied model below the tied one at every shape. At 100B, Wikipedia and C4 agree in direction: every Foil is below Base and every untied model is below its tied counterpart. arXiv is the exception: at 20B Foil-3 and the $(16,4,4)$ untied model are above their comparators, and Foil-3 remains above Base at 100B. Slice sizes 3--8M tokens; the evaluation reports means only, so no standard errors. Code and book slices are absent from the training corpus and are not reported. The slices are small and no standard errors are available, so differences of about 0.001 nat or less may be within noise.}
\label{tab:app-heldout}
\begin{tabular}{@{}l c r r r r r@{}}
\toprule
Model & $(E,D,L)$ & \begin{tabular}[b]{@{}c@{}}Training\\stream\end{tabular} & FineWeb-Edu & Wikipedia & C4 & arXiv \\
\midrule
\multicolumn{7}{@{}l}{\emph{20B tokens}} \\
Base-tied & (8,8,2) & 2.6884 & 2.6307 & 2.6840 & 3.1288 & 2.9151 \\
proto-Foil-3 & (16,4,4) & 2.6793 & 2.6198 & 2.6782 & 3.1190 & 2.9147 \\
proto-Foil-2 & (32,2,8) & 2.6836 & 2.6240 & 2.6933 & 3.1251 & 2.9395 \\
proto-Foil-1 & (64,1,16) & 2.7005 & 2.6414 & 2.7052 & 3.1410 & 2.9733 \\
Base & (8,8,2) & 2.6659 & 2.6077 & 2.6703 & 3.1043 & 2.8997 \\
Foil-3 & (16,4,4) & 2.6627 & 2.6035 & 2.6704 & 3.1021 & 2.9337 \\
Foil-2 & (32,2,8) & 2.6566 & 2.5984 & 2.6651 & 3.0952 & 2.8729 \\
Foil-1 & (64,1,16) & 2.6586 & 2.6000 & 2.6731 & 3.0981 & 2.8765 \\
\midrule
\multicolumn{7}{@{}l}{\emph{100B tokens}} \\
Base-tied & (8,8,2) & 2.5218 & 2.4711 & 2.5408 & 2.9824 & 2.7153 \\
proto-Foil-3 & (16,4,4) & 2.5202 & 2.4700 & 2.5430 & 2.9809 & 2.7447 \\
proto-Foil-2 & (32,2,8) & 2.5305 & 2.4800 & 2.5533 & 2.9902 & 2.7795 \\
proto-Foil-1 & (64,1,16) & 2.5481 & 2.4971 & 2.5632 & 3.0078 & 2.7933 \\
Base & (8,8,2) & 2.5108 & 2.4610 & 2.5376 & 2.9721 & 2.7127 \\
Foil-3 & (16,4,4) & 2.5061 & 2.4552 & 2.5334 & 2.9649 & 2.7231 \\
Foil-2 & (32,2,8) & 2.5012 & 2.4506 & 2.5135 & 2.9594 & 2.7084 \\
Foil-1 & (64,1,16) & 2.4988 & 2.4492 & 2.5311 & 2.9584 & 2.7084 \\
\bottomrule
\end{tabular}
\end{table}

\section{Ablation results}

\paragraph{Ablation details.}
\label{sec:app-ablation}
\Cref{tab:app-interaction} lists the widening gains and the interaction of widening and looping behind \Cref{sec:abl-widen}; \Cref{tab:app-per-pass} lists the routing confidence and the looping gains of the $(8,8,L)$ series behind \Cref{sec:abl-peak}.

\begin{table}[h]
\centering
\caption{Widening and looping, 20B tokens, tied attention. Gains are decreases of the final-window loss (nat), $\pm1$ standard error of the paired difference. Top: widening doubles $E$ at fixed $D$ and $L$; there are no $(4,8,8)$, $(16,8,4)$ or $(16,8,8)$ models. Bottom: for each $2\times2$ block, looping doubles $L$ and ``both'' does the two steps together; interaction = both $-$ widening $-$ looping, with a standard error of about $0.0014$ combined from the independent paired differences.}
\label{tab:app-interaction}
\footnotesize
\begin{tabular}{lccc}
\toprule
Widening & $L=2$ & $L=4$ & $L=8$ \\
\midrule
$(8,4,L)\to(16,4,L)$ & $0.0384\pm0.0011$ & $0.0402\pm0.0010$ & $0.0491\pm0.0012$ \\
$(4,8,L)\to(8,8,L)$ & $0.0393\pm0.0010$ & $0.0460\pm0.0009$ & -- \\
$(8,8,L)\to(16,8,L)$ & $0.0470\pm0.0009$ & -- & -- \\
\bottomrule
\end{tabular}

\vspace{0.6em}
\begin{tabular}{lcccc}
\toprule
Start & widening & looping (passes) & both & interaction \\
\midrule
$(8,4,2)$ & $0.0384$ & $0.0678$ ($2\to4$) & $0.1080$ & $+0.0018$ \\
$(8,4,4)$ & $0.0402$ & $0.0321$ ($4\to8$) & $0.0812$ & $+0.0089$ \\
$(4,8,2)$ & $0.0393$ & $0.0530$ ($2\to4$) & $0.0990$ & $+0.0067$ \\
\bottomrule
\end{tabular}
\end{table}

\begin{table}[h]
\centering
\caption{The $(8,8,L)$ series, 20B tokens, tied attention. \TMCR{}: model-level routing confidence of the recurrent core (geometric mean). Gain: decrease of the final-window loss relative to the non-looped $(8,8,1)$ (nat, $\pm1$ standard error); per pass: gain divided by the $L-1$ added passes.}
\label{tab:app-per-pass}
\footnotesize
\begin{tabular}{lccccc}
\toprule
$L$ & $1$ & $2$ & $3$ & $4$ & $8$ \\
\midrule
\TMCR{} & $4.06$ & $4.15$ & $4.02$ & $3.73$ & $3.36$ \\
Gain & -- & $0.0640\pm0.0013$ & $0.1022\pm0.0013$ & $0.1237\pm0.0014$ & $0.1564\pm0.0014$ \\
Per pass & -- & $0.0640$ & $0.0511$ & $0.0412$ & $0.0223$ \\
\bottomrule
\end{tabular}
\end{table}

\section{Routing metrics of all models}
\begin{table}[h]
\centering
\caption{Routing metrics of the recurrent core for all small-tier models at 20B tokens (step 50,000). Class~0: distinct real experts used per token, its upper bound $D\min(E,kL)$ and the ratio of the two; class~1: model-level $B_2$; class~2: \TMCR{}. $^\S$Repeated from an earlier group. Values depend on the pool size $E$; compare absolute values only at equal shape (\Cref{sec:method-metrics}).}
\label{tab:app-metrics}
\resizebox{\linewidth}{!}{%
\begin{tabular}{lccccccc}
\toprule
Model & $(E,D,L)$ & attention & distinct experts & upper bound & ratio & $B_2$ & \TMCR{} \\
\midrule
\multicolumn{8}{l}{\emph{Flattening, tied attention}} \\
Base-tied & $(8,8,2)$ & tied & $23.47$ & 32 & $0.73$ & $0.921$ & $4.15$ \\
proto-Foil-3 & $(16,4,4)$ & tied & $15.67$ & 32 & $0.49$ & $0.849$ & $5.74$ \\
proto-Foil-2 & $(32,2,8)$ & tied & $11.61$ & 32 & $0.36$ & $0.807$ & $7.46$ \\
proto-Foil-1 & $(64,1,16)$ & tied & $9.42$ & 32 & $0.29$ & $0.731$ & $8.39$ \\
\midrule
\multicolumn{8}{l}{\emph{Flattening, untied attention}} \\
Base & $(8,8,2)$ & untied & $24.08$ & 32 & $0.75$ & $0.947$ & $4.21$ \\
Foil-3 & $(16,4,4)$ & untied & $17.79$ & 32 & $0.56$ & $0.898$ & $6.14$ \\
Foil-2 & $(32,2,8)$ & untied & $12.48$ & 32 & $0.39$ & $0.839$ & $8.50$ \\
Foil-1 & $(64,1,16)$ & untied & $9.71$ & 32 & $0.30$ & $0.805$ & $8.77$ \\
\midrule
\multicolumn{8}{l}{\emph{Loop passes, $(E,D)=(8,8)$}} \\
-- & $(8,8,1)$ & tied & $16.00$ & 16 & -- & $0.952$ & $4.06$ \\
Base-tied$^\S$ & $(8,8,2)$ & tied & $23.47$ & 32 & $0.73$ & $0.921$ & $4.15$ \\
-- & $(8,8,3)$ & tied & $26.34$ & 48 & $0.55$ & $0.912$ & $4.02$ \\
-- & $(8,8,4)$ & tied & $28.44$ & 64 & $0.44$ & $0.915$ & $3.73$ \\
-- & $(8,8,8)$ & tied & $36.23$ & 64 & $0.57$ & $0.908$ & $3.36$ \\
\midrule
\multicolumn{8}{l}{\emph{Loop passes, $(E,D)=(16,4)$}} \\
-- & $(16,4,2)$ & tied & $12.22$ & 16 & $0.76$ & $0.897$ & $6.08$ \\
proto-Foil-3$^\S$ & $(16,4,4)$ & tied & $15.67$ & 32 & $0.49$ & $0.849$ & $5.74$ \\
-- & $(16,4,6)$ & tied & $18.66$ & 48 & $0.39$ & $0.830$ & $5.34$ \\
-- & $(16,4,8)$ & tied & $21.21$ & 64 & $0.33$ & $0.838$ & $5.26$ \\
\midrule
\multicolumn{8}{l}{\emph{Loop passes, $(E,D)=(8,4)$}} \\
-- & $(8,4,2)$ & tied & $11.53$ & 16 & $0.72$ & $0.941$ & $4.19$ \\
-- & $(8,4,4)$ & tied & $13.57$ & 32 & $0.42$ & $0.932$ & $3.80$ \\
-- & $(8,4,8)$ & tied & $16.99$ & 32 & $0.53$ & $0.904$ & $3.53$ \\
\midrule
\multicolumn{8}{l}{\emph{Pool size and depth at fixed $\Ecomp=32$}} \\
-- & $(4,8,2)$ & tied & $20.73$ & 32 & $0.65$ & $0.969$ & $2.06$ \\
Base-tied$^\S$ & $(8,8,2)$ & tied & $23.47$ & 32 & $0.73$ & $0.921$ & $4.15$ \\
-- & $(16,8,2)$ & tied & $24.76$ & 32 & $0.77$ & $0.903$ & $6.07$ \\
--$^\S$ & $(8,4,4)$ & tied & $13.57$ & 32 & $0.42$ & $0.932$ & $3.80$ \\
-- & $(8,16,1)$ & tied & $32.00$ & 32 & -- & $0.926$ & $3.84$ \\
\midrule
\multicolumn{8}{l}{\emph{Other controls}} \\
-- & $(4,8,4)$ & tied & $23.27$ & 32 & $0.73$ & $0.963$ & $1.88$ \\
Base-tied, two-layer prelude/coda & $(8,8,2)$ & tied & $22.54$ & 32 & $0.70$ & $0.912$ & $4.26$ \\
Base-tied, seed 43 & $(8,8,2)$ & tied & $23.30$ & 32 & $0.73$ & $0.923$ & $4.34$ \\
\bottomrule
\end{tabular}}
\end{table}

\begin{table}[t]
\centering
\footnotesize
\setlength{\tabcolsep}{3pt}
\caption{Values of related-work routing metrics for every base-width ($d_{\text{model}}=1024$) model at the end of 20B-token training (step 50{,}000). Loop block only. Load metrics: $B_H$ = normalised load entropy $H(q)/\log E$; MaxVio $= E\max_i q_i - 1$; $G^{*}$ = Gini coefficient normalised by $E-1$; each is computed per physical layer from selection counts pooled over all passes, then averaged arithmetically over physical layers. Confidence metrics (per token, then averaged over tokens): rank-1 mean probability; selected-set mass $M_k$ (sum of the $k$ selected full-pool probabilities); boundary margin (logit of the $k$-th minus the $(k{+}1)$-th expert); $C_{\text{full}} = 1 - H(p)/\log E$; $C_{\text{sel}}$ = one minus the normalised entropy of the renormalised weights of the selected experts; each is averaged with equal weight over all (physical layer, pass) cells of the loop block (arithmetic mean). Selections are recomputed as the top-$k$ of the stored float16 router logits. Token drop rate is 0 for every model and not applicable (no capacity limit, no dropping), so it is not listed. Probe set: $255{,}500$ tokens from $500$ documents, $50$ from each of ten domains of the Pile. Model names and footnotes as in \Cref{tab:app-hparams}.}
\label{tab:app-related-values}
\begin{tabular}{@{}l l c r r r r r r r r@{}}
\toprule
 & & & \multicolumn{3}{c}{Load} & \multicolumn{5}{c}{Confidence} \\
\cmidrule(lr){4-6}\cmidrule(l){7-11}
Model & Run id & $(E,D,L)$ & $B_H$ & MaxVio & $G^{*}$ & \begin{tabular}[b]{@{}c@{}}Rank-1\\prob.\end{tabular} & \begin{tabular}[b]{@{}c@{}}Selected\\mass $M_k$\end{tabular} & \begin{tabular}[b]{@{}c@{}}Boundary\\margin\end{tabular} & $C_{\text{full}}$ & $C_{\text{sel}}$ \\
\midrule
\multicolumn{11}{@{}l}{\emph{Foil family}} \\
Base-tied & S1 & (8,8,2) & 0.979 & 0.523 & 0.162 & 0.382 & 0.556 & 0.417 & 0.164 & 0.130 \\
proto-Foil-3 & S2 & (16,4,4) & 0.971 & 1.099 & 0.221 & 0.269 & 0.403 & 0.374 & 0.143 & 0.103 \\
proto-Foil-2 & S3 & (32,2,8) & 0.972 & 1.812 & 0.236 & 0.173 & 0.274 & 0.295 & 0.130 & 0.069 \\
proto-Foil-1 & S4 & (64,1,16) & 0.968 & 3.449 & 0.270 & 0.095 & 0.159 & 0.198 & 0.110 & 0.040 \\
Base & U1 & (8,8,2) & 0.987 & 0.451 & 0.138 & 0.386 & 0.559 & 0.409 & 0.168 & 0.133 \\
Foil-3 & U2 & (16,4,4) & 0.982 & 0.964 & 0.176 & 0.286 & 0.411 & 0.359 & 0.150 & 0.133 \\
Foil-2 & U3 & (32,2,8) & 0.977 & 1.479 & 0.216 & 0.193 & 0.294 & 0.330 & 0.137 & 0.092 \\
Foil-1 & U4 & (64,1,16) & 0.975 & 1.291 & 0.249 & 0.106 & 0.171 & 0.229 & 0.106 & 0.055 \\
\addlinespace
\multicolumn{11}{@{}l}{\emph{Other grid models}} \\
-- & S6$^{c}$ & (8,8,1) & 0.987 & 0.337 & 0.126 & 0.381 & 0.561 & 0.437 & 0.161 & 0.126 \\
-- & S8 & (8,8,3) & 0.977 & 0.589 & 0.172 & 0.375 & 0.553 & 0.406 & 0.159 & 0.121 \\
-- & S9 & (8,8,4) & 0.979 & 0.644 & 0.178 & 0.361 & 0.543 & 0.391 & 0.148 & 0.106 \\
-- & N4 & (8,8,8) & 0.977 & 0.651 & 0.191 & 0.343 & 0.531 & 0.363 & 0.140 & 0.086 \\
-- & S5 & (16,4,2) & 0.980 & 0.759 & 0.178 & 0.288 & 0.412 & 0.396 & 0.149 & 0.139 \\
-- & N3 & (16,4,6) & 0.968 & 1.366 & 0.229 & 0.253 & 0.390 & 0.336 & 0.138 & 0.084 \\
-- & N2 & (16,4,8) & 0.971 & 1.339 & 0.214 & 0.250 & 0.389 & 0.315 & 0.141 & 0.078 \\
-- & S7 & (8,4,2) & 0.985 & 0.409 & 0.141 & 0.383 & 0.564 & 0.452 & 0.164 & 0.123 \\
-- & S12 & (8,4,4) & 0.983 & 0.517 & 0.156 & 0.365 & 0.551 & 0.413 & 0.151 & 0.103 \\
-- & N1 & (8,4,8) & 0.977 & 0.712 & 0.186 & 0.352 & 0.552 & 0.405 & 0.158 & 0.078 \\
-- & S11$^{a}$ & (4,8,2) & 0.989 & 0.253 & 0.120 & 0.493 & 0.739 & 0.502 & 0.153 & 0.116 \\
-- & S13$^{a}$ & (4,8,4) & 0.985 & 0.255 & 0.126 & 0.473 & 0.728 & 0.488 & 0.133 & 0.094 \\
-- & S10 & (16,8,2) & 0.981 & 0.779 & 0.175 & 0.288 & 0.410 & 0.355 & 0.152 & 0.141 \\
-- & S14$^{c}$ & (8,16,1) & 0.981 & 0.489 & 0.160 & 0.368 & 0.549 & 0.411 & 0.151 & 0.113 \\
-- & S15$^{b}$ & (8,8,2) & 0.977 & 0.583 & 0.172 & 0.388 & 0.564 & 0.418 & 0.170 & 0.132 \\
\bottomrule
\end{tabular}
\end{table}

\section{Expert-masking evaluation}
\label{sec:app-masking}

The masking test asks whether experts that receive little traffic are also of little value. It uses two disjoint samples of the same corpus: sample A selects the experts, sample B measures the loss. On sample A we count, for every physical expert of the recurrent core, how often it is selected into the top-$k$, with the passes pooled onto the physical expert; the prelude and coda layers are never masked. Sorting the experts by this share from lowest to highest and adding them until the cumulative share is closest to $x\%$ of all routing requests gives the least-used group $T(x)$. Masking sets the routing probabilities of the masked experts to zero before the top-$k$ selection, so that each token chooses its $k$ experts among the remaining ones with renormalised weights; nothing is retrained. We report the increase $\Delta L_T$ of the mean language-modelling loss on sample B, and compare it with 20 random groups drawn from the experts outside $T(x)$ and matched to its traffic share (fixed seeds). Every masked set must leave at least $k$ experts in each layer. We use $x=10\%$ (\Cref{tab:app-masking}).

\begin{figure}[h]
\begin{center}
\includegraphics[width=\linewidth]{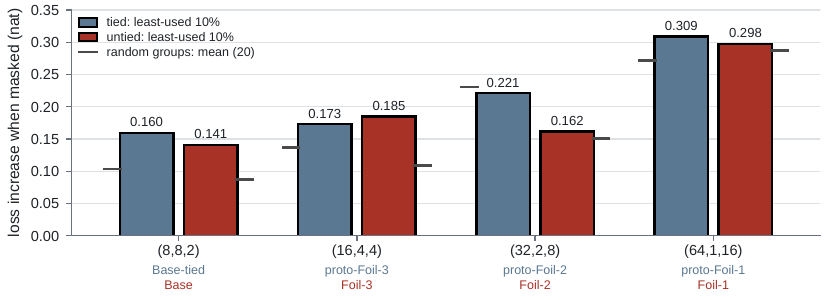}
\end{center}
\caption{Masking the least-used experts at the 10\% traffic tier, 20B tokens. For each shape, the left bar is the tied-attention model and the right bar the untied-attention model; bar height is the loss increase from masking the least-used group and the short dark dash beside it the mean loss increase of the 20 traffic-matched random groups. The ranges of the random groups are given in \Cref{tab:app-masking}, the ratios to the random mean in \Cref{tab:app-masking-ratio}.}
\label{fig:app-masking-10}
\end{figure}

At the 10\% tier, across all 16 combinations of the eight models and the two token budgets, the loss increase from masking the least-used group never falls below the range of the random groups: rarely used experts are not dispensable. On the least flattened models at 20B tokens the least-used group costs more than almost every random group (19 or 20 of 20 for Base-tied, Base and Foil-3; \Cref{tab:app-masking,tab:app-masking-ratio} and \Cref{fig:app-masking-10}). Low load therefore does not indicate low value, and an unbalanced load is not by itself a sign of an unhealthy router.

\begin{table}[h]
\centering
\caption{Masking the least-used experts at the 10\% traffic tier. $\Delta L_T$: loss increase (nat) when the least-used group is masked; random: loss increase for 20 traffic-matched random groups (mean and range); last column: number of the 20 random groups whose loss increase is below $\Delta L_T$.}
\label{tab:app-masking}
\footnotesize
\begin{tabular}{llccccc}
\toprule
Model & $(E,D,L)$ & experts masked & $\Delta L_T$ & random mean & random range & above random \\
\midrule
\multicolumn{7}{l}{\emph{20B tokens}} \\
Base-tied & (8,8,2) & 11 & $0.160$ & $0.104$ & $[0.068,\,0.165]$ & 19/20 \\
proto-Foil-3 & (16,4,4) & 11 & $0.173$ & $0.137$ & $[0.047,\,0.229]$ & 15/20 \\
proto-Foil-2 & (32,2,8) & 11 & $0.221$ & $0.231$ & $[0.129,\,0.386]$ & 9/20 \\
proto-Foil-1 & (64,1,16) & 12 & $0.309$ & $0.271$ & $[0.140,\,0.464]$ & 15/20 \\
Base & (8,8,2) & 9 & $0.141$ & $0.087$ & $[0.044,\,0.143]$ & 19/20 \\
Foil-3 & (16,4,4) & 10 & $0.185$ & $0.109$ & $[0.066,\,0.141]$ & 20/20 \\
Foil-2 & (32,2,8) & 10 & $0.162$ & $0.151$ & $[0.087,\,0.235]$ & 12/20 \\
Foil-1 & (64,1,16) & 10 & $0.298$ & $0.287$ & $[0.099,\,0.779]$ & 15/20 \\
\midrule
\multicolumn{7}{l}{\emph{100B tokens}} \\
Base-tied & (8,8,2) & 10 & $0.109$ & $0.121$ & $[0.048,\,0.461]$ & 15/20 \\
proto-Foil-3 & (16,4,4) & 11 & $0.174$ & $0.201$ & $[0.092,\,0.508]$ & 9/20 \\
proto-Foil-2 & (32,2,8) & 11 & $0.231$ & $0.239$ & $[0.091,\,0.426]$ & 11/20 \\
proto-Foil-1 & (64,1,16) & 12 & $0.232$ & $0.189$ & $[0.113,\,0.390]$ & 16/20 \\
Base & (8,8,2) & 9 & $0.134$ & $0.113$ & $[0.061,\,0.177]$ & 12/20 \\
Foil-3 & (16,4,4) & 10 & $0.131$ & $0.196$ & $[0.069,\,0.693]$ & 13/20 \\
Foil-2 & (32,2,8) & 11 & $0.136$ & $0.236$ & $[0.096,\,2.035]$ & 11/20 \\
Foil-1 & (64,1,16) & 11 & $0.235$ & $0.269$ & $[0.078,\,1.223]$ & 16/20 \\
\bottomrule
\end{tabular}
\end{table}

\begin{table}[h]
\centering
\caption{Least-used group versus random groups at the 10\% traffic tier, 20B tokens. Ratio: loss increase from masking the least-used group divided by the mean loss increase of the 20 traffic-matched random groups; last column: number of random groups whose loss increase is below that of the least-used group. Seven of the eight ratios exceed one; none of the least-used groups falls below the range of the random groups.}
\label{tab:app-masking-ratio}
\footnotesize
\begin{tabular}{llccc}
\toprule
Model & $(E,D,L)$ & attention & ratio & random groups below \\
\midrule
Base-tied & $(8,8,2)$ & tied & $1.54$ & 19/20 \\
proto-Foil-3 & $(16,4,4)$ & tied & $1.26$ & 15/20 \\
proto-Foil-2 & $(32,2,8)$ & tied & $0.96$ & 9/20 \\
proto-Foil-1 & $(64,1,16)$ & tied & $1.14$ & 15/20 \\
Base & $(8,8,2)$ & untied & $1.61$ & 19/20 \\
Foil-3 & $(16,4,4)$ & untied & $1.69$ & 20/20 \\
Foil-2 & $(32,2,8)$ & untied & $1.07$ & 12/20 \\
Foil-1 & $(64,1,16)$ & untied & $1.04$ & 15/20 \\
\bottomrule
\end{tabular}
\end{table}

\section{Limitations}
\label{sec:app-limitations}
Four limitations bound our conclusions. First, we have no non-looped control at equal parameters and compute: the non-looped model with the expert parameters and expert calls of $(8,8,2)$ is $(4,16,1)$, which raises the active share $k/E$ to $1/2$ and so confounds sharing experts across layers with the sparsity of activation. Our question is how the experts should be arranged within the looped block once looping has been chosen; that sharing experts across layers is itself useful is supported by external evidence~\citep{qiu2026more, jaggi2026tying}. Second, apart from one seed replicate, every model is trained with a single seed; the replicate differs by $-0.0003\pm0.0009$~nat, which sets the resolution of our comparisons, every difference interpreted in the main text is at least $0.002$~nat, and the loss standard errors are optimistic because consecutive steps are not independent (Appendix~\ref{sec:app-training}). Third, all results are at a single width, $\dmodel=1024$. Fourth, the ablations use shared attention, under which the gain from flattening vanishes after the first step (\Cref{fig:untied}, a1--a2); untying the attention is what lets the gain continue, so the ablation trends may differ in size for Foil itself.

\end{document}